\documentclass[sigconf,nonacm,pdfa]{acmart}

\usepackage[utf8]{inputenc}
\usepackage{amsmath,amsfonts}
\usepackage{amsthm}
\usepackage{textcomp}
\usepackage[english]{babel}
\usepackage[nounderscore]{syntax}
\usepackage{makecell,multirow}
\usepackage{color}
\usepackage[table]{xcolor}
\usepackage{array}
\usepackage{enumitem}
\usepackage{epstopdf}
\usepackage{booktabs}
\usepackage{threeparttable}
\usepackage{bm}
\usepackage{graphicx}
\usepackage{subfigure}
\usepackage{caption}
\usepackage{balance}
\usepackage{svg}
\usepackage{algpseudocode}
\usepackage{amsmath}
\usepackage{longtable}
\usepackage{graphicx}
\usepackage{stfloats}
\usepackage{placeins}

\usepackage{siunitx}

\newtheoremstyle{mydefn}
{}{}
{\it}       % body font
{0pt}       % indent
{\bfseries} % head font
{:~}        % punctuation after head
{0.25em}    % spacing after head
{}          % CUSTOM-HEAD-SPEC
\theoremstyle{mydefn}

\newtheorem{definition}{Definition}[section]

\newtheoremstyle{myexample}
{}{}
{}          % body font
{0pt}       % indent
{\bfseries} % head font
{:~}        % punctuation after head
{0.25em}    % spacing after head
{}          % CUSTOM-HEAD-SPEC
\theoremstyle{myexample}

\newtheorem{example}{Example}[section]

\renewcommand{\paragraph}[1]{\smallskip\noindent\textbf{#1.}}
\renewcommand{\subparagraph}[1]{\smallskip\noindent\textbf{\underline{#1.}}}
\renewcommand{\sec}[1]{\textbf{\S#1}}

\newcommand{\floor}[1]{\lfloor #1 \rfloor}

\makeatletter
\newif\if@restonecol
\makeatother

\usepackage[linesnumbered,ruled,vlined]{algorithm2e}
\usepackage{algpseudocode}

\SetKwComment{Comment}{$\triangleright$\ }{}
\SetKwRepeat{Do}{do}{while}

\newcommand\vldbdoi{XX.XX/XXX.XX}
\newcommand\vldbpages{XXX-XXX}
\newcommand\vldbyear{2020}
\newcommand\vldbauthors{\authors}
\newcommand\vldbtitle{\shorttitle} 
\newcommand\vldbpagestyle{plain} 

\begin{document}

\title{RFOD: Random Forest-based Outlier Detection for Tabular Data}

\author{Yihao Ang}
\orcid{0009-0009-1564-4937}
\affiliation{
 \institution{National University of Singapore}
 \country{}
}
\email{yihao\_ang@comp.nus.edu.sg}

\author{Peicheng Yao}
\orcid{0000-0003-0000-xxxx}
\affiliation{
 \institution{National University of Singapore}
 \country{}
}
\email{peicheng.yao@u.nus.edu}

\author{Yifan Bao}
\orcid{0009-0000-9672-0747}
\affiliation{
 \institution{National University of Singapore}
 \country{}
}
\email{yifan\_bao@comp.nus.edu.sg}

\author{Yushuo Feng}
\orcid{0009-0000-3280-9881}
\authornote{Work done during internship at the National University of Singapore.}
\affiliation{
 \institution{Huazhong University of Science \& Technology}
 \country{}
}
\email{u202315134@hust.edu.cn}
% \footnote{Work done during internship at the National University of Singapore}

\author{Qiang Huang}
% \authornote{Qiang Huang is the corresponding author.}
\orcid{0000-0003-1120-4685}
\affiliation{
 \institution{Harbin Institute of Technology (Shenzhen)}
 \country{}
}
\email{huangqiang@hit.edu.cn}

\author{Anthony K. H. Tung}
\orcid{0000-0001-7300-6196}
\affiliation{
 \institution{National University of Singapore}
 \country{}
}
\email{atung@comp.nus.edu.sg}

\author{Zhiyong Huang}
\orcid{0000-0002-1931-7775}
\affiliation{
 \institution{National University of Singapore}
 \country{}
}
\email{huangzy@comp.nus.edu.sg}

\begin{abstract}
Outlier detection in tabular data is crucial for safeguarding data integrity in high-stakes domains such as cybersecurity, financial fraud detection, and healthcare, where anomalies can cause serious operational and economic impacts.
Despite advances in both data mining and deep learning, many existing methods struggle with mixed-type tabular data, often relying on encoding schemes that lose important semantic information. 
Moreover, they frequently lack interpretability, offering little insight into which specific values cause anomalies.
To overcome these challenges, we introduce \textsf{\textbf{RFOD}}, a novel \textsf{\textbf{R}}andom \textsf{\textbf{F}}orest-based \textsf{\textbf{O}}utlier \textsf{\textbf{D}}etection framework tailored for tabular data. 
Rather than modeling a global joint distribution, \textsf{RFOD} reframes anomaly detection as a feature-wise conditional reconstruction problem, training dedicated random forests for each feature conditioned on the others. 
This design robustly handles heterogeneous data types while preserving the semantic integrity of categorical features.
To further enable precise and interpretable detection, \textsf{RFOD} combines Adjusted Gower's Distance (AGD) for cell-level scoring, which adapts to skewed numerical data and accounts for categorical confidence, with Uncertainty-Weighted Averaging (UWA) to aggregate cell-level scores into robust row-level anomaly scores.
Extensive experiments on 15 real-world datasets demonstrate that \textsf{RFOD} consistently outperforms state-of-the-art baselines in detection accuracy while offering superior robustness, scalability, and interpretability for mixed-type tabular data.
\end{abstract}

\maketitle

\pagestyle{\vldbpagestyle}
\begingroup\small\noindent\raggedright\textbf{ACM Reference Format:}\\
\vldbauthors. \vldbtitle. ACM Conference, \vldbpages, \vldbyear.\\
\href{https://doi.org/\vldbdoi}{doi:\vldbdoi}
\endgroup
\begingroup
\renewcommand\thefootnote{}\footnote{\noindent
Permission to make digital or hard copies of all or part of this work for personal or classroom use is granted without fee provided that copies are not made or distributed for profit or commercial advantage and that copies bear this notice and the full citation on the first page. Copyrights for components of this work owned by others than ACM must be honored. Abstracting with credit is permitted. To copy otherwise, or republish, to post on servers or to redistribute to lists, requires prior specific permission and/or a fee. Request permissions from permissions@acm.org. \\ 
\raggedright ACM Conference, ISSN XXXX-XXXX. \\
\href{https://doi.org/\vldbdoi}{doi:\vldbdoi} \\
}\addtocounter{footnote}{-1}\endgroup

\section{Introduction}
\label{sect:intro}

%%%
Outliers, data instances whose feature combinations significantly deviate from the dominant distribution, pose substantial challenges for data-driven decision-making \cite{aggarwal2017outlier}. 
Undetected anomalies can distort analyses, conceal critical insights, and disrupt operations. 
These risks are particularly acute in high-stakes domains such as cybersecurity \cite{moustafa2015unsw, moustafa2016evaluation}, financial fraud detection \cite{cao2018collective, tao2019mvan}, healthcare \cite{smith1988using, cardiotocography_193}, and industrial monitoring \cite{cad, ang2024eads, bao2024towards}, where timely and interpretable anomaly detection is essential for ensuring reliability and economic viability.

%%%
%%%
In tabular data, effective outlier detection approaches require satisfying two interlinked goals: 
(1) accurately identify anomalous instances, and (2) pinpoint the specific feature values responsible for the anomalies. 
Achieving both objectives is crucial not only for high detection performance but also for providing interpretable and actionable insights necessary in real-world applications.
%%%
Despite extensive research \cite{aggarwal2017outlier, kriegel2008angle, li2020copod, hardin2004outlier, goldstein2012histogram, pevny2016loda, zhao2019lscp, bandaragoda2018isolation, liu2021rca, pang2018learning, wang2021unsupervised, bergman2020classification, qiu2021neural}, outlier detection on tabular data remains particularly challenging.
Unlike homogeneous data types such as images or text, tabular datasets often consist of mixed numerical and categorical features with diverse distributions and complex interdependencies \cite{han2022adbench, cortes2020explainable, Ang2023TSGBench, ang2025ctbench}. 
Moreover, anomalies frequently affect only specific cells within a row, rendering purely row-level methods insufficient for precise localization \cite{akrami2022robust}.

%%%
Vanilla data mining approaches, such as Local Outlier Factor (\textsf{LOF}) \cite{breunig2000lof}, Isolation Forest (\textsf{IF}) \cite{liu2008if}, One-Class Support Vector Machines (\textsf{OCSVM}) \cite{wang2019osvm}, Empirical Cumulative Distribution (\textsf{ECOD}) \cite{li2022ecod}, and OutlierTree (\textsf{OT}) \cite{cortes2020explainable}, typically rely on global proximity or statistical heuristics.
While efficient and easy to implement, they often treat features independently and fail to capture the subtle multivariate relationships critical for detecting contextual anomalies.
%%%
In contrast, modern deep learning methods, including Deep Support Vector Data Description (\textsf{Deep SVDD}) \cite{deepsvdd}, Deviation Networks (\textsf{DevNet}) \cite{devnet}, Internal Contrastive Learning (\textsf{ICL}) \cite{icl}, Scale Learning-based deep Anomaly Detection (\textsf{SLAD}) \cite{slad}, and Deep Isolation Forest (\textsf{DIF}) \cite{dif}, attempt to learn complex dependencies through latent representations. 
%%%
Nevertheless, these approaches predominantly assume purely numerical inputs and suffer from two critical limitations: 
(1) they are not natively designed for \emph{mixed-type} data and rely on preprocessing (e.g., one-hot encoding) that may lose semantic nuance, and 
(2) their \emph{black-box} nature impedes interpretability, making it difficult to trace the rationale behind anomaly scores.

\begin{figure*}[t]
  \centering
  \includegraphics[width=0.99\textwidth]{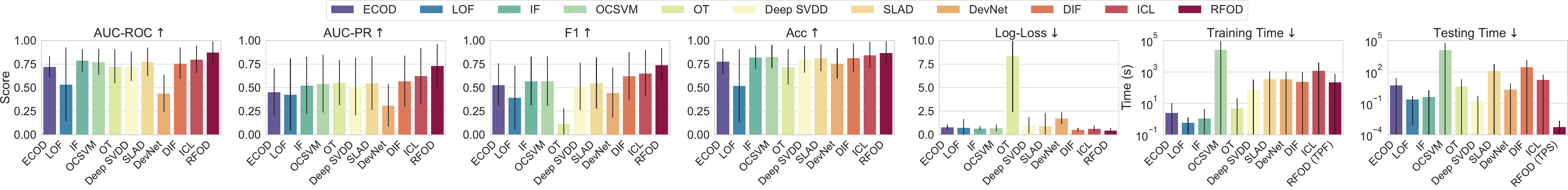}
  \vspace{-1.0em}
  \caption{Average anomaly detection performance of all methods across 15 real-world tabular datasets. \textsf{RFOD} consistently ranks at or near the top, demonstrating strong accuracy and robustness over both data mining and deep learning baselines.} 
  \label{fig:main_results}
  \vspace{-0.75em}
\end{figure*}

%%%
Figure \ref{fig:main_results} from our experiments highlights these challenges. 
%%%
While some deep learning methods occasionally excel on specific datasets, their performance is inconsistent, particularly for mixed-type data.
%%%
Traditional data mining methods, though faster and more interpretable, often lag in detection accuracy and fine-grained insights.
No existing method consistently delivers strong performance across the diverse landscape of tabular data.

%%%
To bridge this gap, we introduce \textsf{\textbf{RFOD}}, a novel \textsf{\textbf{R}}andom \textsf{\textbf{F}}orest-based \textsf{\textbf{O}}utlier \textsf{\textbf{D}}etection framework specifically designed for tabular data. 
%%%
Unlike conventional methods that model the joint distribution of all features, RFOD reframes anomaly detection as a \textbf{feature-wise conditional reconstruction} problem.
%%%
It trains dedicated random forests for each feature, conditioned on the remaining features, enabling nuanced modeling of inter-feature dependencies and flexible adaptation to heterogeneous data types. 
During inference, \textsf{RFOD} reconstructs each feature value for test instances, quantifies deviations via residuals, adjusts anomaly scores based on model uncertainty, and aggregates these signals into robust row-level anomaly scores.
%%%
\textsf{RFOD} distinguishes itself from existing methods through several key innovations, each designed to address the challenges of mixed-type tabular data:
\begin{itemize}[nolistsep,leftmargin=20pt]
  \item \textbf{Feature-Specific Random Forests:} 
  Independently trained forests for each feature capture localized conditional relationships, bypassing restrictive global assumptions and enabling precise localization of anomaly signals, particularly when anomalies affect only a subset of features.

  \item \textbf{Forest Pruning:} 
  A principled mechanism using out-of-bag (OOB) validation to retain only the most informative decision trees in each forest, reducing noise and improving both generalization and computational efficiency.

  \item \textbf{Adjusted Gower's Distance (AGD):} 
  We introduce AGD for fine-grained, cell-level anomaly measurement. 
  Unlike traditional distance metrics, AGD adapts to skewed numerical distributions via quantile normalization and incorporates confidence-weighted matching for categorical features, yielding robust, interpretable cell-level anomaly scores.

  \item \textbf{Uncertainty-Weighted Averaging (UWA):} 
  \textsf{RFOD} aggregates cell-level anomaly scores into row-level scores through UWA, dynamically weighting cell contributions based on prediction confidence. 
  This mechanism prevents noisy or uncertain predictions from diluting the overall anomaly signal. 
\end{itemize}

%%%
Our comprehensive experiments across 15 real-world tabular datasets demonstrate \textsf{RFOD}'s substantial empirical advantages.
%%% 
As shown in Figure~\ref{fig:main_results}, \textsf{RFOD} consistently achieves the highest average anomaly detection performance across all key metrics, including AUC-ROC, AUC-PR, F1, Accuracy, and Log-Loss, surpassing ten state-of-the-art baselines spanning both data mining and deep learning paradigms.
Remarkably, \textsf{RFOD} delivers up to a 9.1\% gain in AUC-ROC over the best competing method and achieves an average 91.2\% reduction in test-time latency, compared to leading alternatives, underscoring its scalability and practical value.
%%%
Ablation studies confirm the critical roles played by \textsf{RFOD}'s components, while detailed case analyses demonstrate how \textsf{RFOD} provides actionable, transparent explanations by pinpointing anomalies at both the cell and row levels.
%%%
Together, these results position \textsf{RFOD} as a robust, interpretable, and scalable solution for outlier detection in tabular data, effectively addressing longstanding gaps in handling mixed-type datasets and delivering insights vital for real-world decision-making.

%%%
The remainder of this paper is structured as follows: Section~\ref{sect:related} reviews related work; Section~\ref{sect:problem} formulates the problem; Section~\ref{sect:methods} details the \textsf{RFOD} framework; Section~\ref{sect:expt} presents extensive experimental results and analyses; and Section~\ref{sect:conclusion} concludes the paper.

\section{Related Work}
\label{sect:related}

Outlier detection for tabular data has long been a central topic in data mining and machine learning. 
Existing research broadly falls into two categories: (1) data mining–based methods and (2) deep learning–based approaches. 
We review each and position our proposed \textsf{RFOD} framework within this landscape.

\subsection{Data Mining-based Outlier Detection}
\label{sect:related:mining}

Traditional data mining methods detect anomalies by measuring deviations from expected patterns using statistical, density-based, or proximity-based criteria. 
These approaches often rely on handcrafted features, pairwise distances, or tree-based models to quantify abnormality \cite{chandola2009adsurvey}.
%%%
%%%
A seminal method in this category is \textsf{IF} \cite{liu2008if}, which isolates outliers through recursive random partitioning, using the average path length of instances in trees as an indicator of anomaly.
%%%
Similarly, \textsf{LOF} \cite{breunig2000lof} assesses how much a data point deviates from the local density of its neighbors, making it effective in heterogeneous density settings.
\textsf{OCSVM} \cite{wang2019osvm} models the decision boundary around normal data in feature space, treating points outside this boundary as anomalies.
%%%
More recently, \textsf{ECOD} \cite{li2022ecod} proposes a nonparametric approach to estimate anomaly scores directly from empirical distributions, focusing on tail probabilities to capture rare events.
%%%
Tree-based methods have also emerged for interpretable detection. 
\textsf{OT} \cite{cortes2020explainable} constructs decision trees that model each feature as a function of others, flagging data points that deviate significantly from expected branches.

Despite their efficiency and interpretability, these methods often suffer from three key limitations:
(1) They usually assume independent feature contributions, overlooking multivariate interactions critical for detecting contextual anomalies.
(2) Many struggle to handle mixed-type data without resorting to potentially lossy encoding schemes.
(3) Their ability to pinpoint cell-level anomalies is limited, hindering detailed diagnostics.
%%%
%%%
\textsf{RFOD} advances this line of research by leveraging Random Forests in a feature-wise conditional reconstruction paradigm. This design offers three advantages:
(1) native support for mixed-type data, eliminating the need for encoding schemes that risk losing feature semantics;
(2) enhanced interpretability through tree-based inference paths, enabling clear attribution of anomalies to specific features and values; and
(3) robustness and generalization, achieved through ensemble averaging and refined further by selective tree pruning, which removes noisy or redundant predictors.

\subsection{Deep-Learning-based Outlier Detection}
\label{sect:related:learning}

The rise of deep learning has fueled a new generation of outlier detection methods that automatically learn complex representations from raw data \cite{dlad}.
These approaches have achieved impressive results, particularly in high-dimensional or structured domains like images and text.
%%%
%%%
For instance, \textsf{Deep SVDD} \cite{deepsvdd} extends traditional Support Vector Data Description (SVDD) \cite{tax2004support} by learning a hypersphere in a latent space to enclose normal samples, flagging deviations as anomalies. 
%%%
\textsf{DevNet} \cite{devnet} introduces a semi-supervised framework that explicitly models deviations between known anomalies and normal instances, improving detection under limited anomaly labels.
%%%
\textsf{ICL} \cite{icl} leverages self-supervised contrastive signals to distinguish anomalies by emphasizing internal dissimilarities in learned representations.
%%%
More recently, \textsf{SLAD} \cite{slad} focuses on learning scale-sensitive transformations, uncovering subtle patterns indicative of anomalies.
%%%
Hybrid approaches have also gained research attention.
\textsf{DIF} \cite{dif} augments traditional \textsf{IF} with neural embeddings to combine expressive power with tree-based interpretability.

However, deep learning–based techniques face significant limitations when applied to tabular data.
%%%
First, most models are designed for numerical inputs and struggle with categorical features, requiring preprocessing (e.g., one-hot encoding or learned embeddings) that can inflate dimensionality or obscure semantic meaning \cite{han2022adbench}.
%%%
Second, neural networks often operate as black boxes, offering limited insight into which features drive anomaly decisions, a critical drawback in domains where explanations are mandatory for trust and compliance \cite{li2022ecod}.
%%%
%%%
\textsf{RFOD} bridges these gaps by integrating the strengths of tree-based approaches with a novel feature-wise reconstruction strategy.
Its design handles mixed-type data seamlessly without complex embeddings and offers principled mechanisms like Adjusted Gower's Distance (AGD) and Uncertainty-Weighted Averaging (UWA) to enhance robustness and interoperability, addressing longstanding challenges in tabular anomaly detection.

\section{Problem Formulation}
\label{sect:problem}

\begin{table}[t]
\centering
% \small
\setlength\tabcolsep{3pt}
\renewcommand{\arraystretch}{1.3}
\caption{Frequently used notations.}
\label{tab:notation}
\vspace{-1.0em}
\resizebox{0.99\columnwidth}{!}{
\begin{tabular}{ll}
    \toprule
    \rowcolor[HTML]{FFF2CC}
    \textbf{Symbol} & \textbf{Description} \\
    \midrule
    $\bm{X}_{\text{train}} \in \mathbb{R}^{n \times d}$  & Training set with $n$ samples and $d$ features: $\bm{X}_{\text{train}} = [x_{i,j}]$ \\
    $\bm{x}_i \in \mathbb{R}^{d}$ & $i$-th sample: $\bm{x}_i = [x_{i,1}, \cdots, x_{i,d}]$ \\
    $\bm{x}^j \in \mathbb{R}^{n}$ & $j$-th feature: $\bm{x}^j = [x_{1,j}, \cdots, x_{n,j}]$ \\
    %%%
    $\bm{RF}_j$ & Random Forest trained for feature $\bm{x}^j$ \\
    $\beta$ & Retaining ratio for forest pruning \\
    %%%
    $\bm{X}_{\text{test}} \in \mathbb{R}^{m \times d}$ & Test set with $m$ samples and $d$ features: $\bm{X}_{\text{test}} = [x_{i,j}]$ \\
    $\bm{\hat{X}}_{\text{test}} \in \mathbb{R}^{m \times d}$ & Reconstructed test set: $\bm{\hat{X}}_{\text{test}} = [\hat{x}_{i,j}]$ \\
    $AGD^{(\text{num})}(x_{i,j}, \hat{x}_{i,j})$ & Adjusted Gower's Distance (AGD) for numerical features \\
    $AGD^{(\text{cat})}(x_{i,j}, \hat{x}_{i,j})$ & Adjusted Gower's Distance (AGD) for categorical features \\
    $\alpha$ & Quantile parameter for AGD for numerical features \\
    $\bm{S}_{\text{cell}} \in \mathbb{R}^{m \times d}$ & Cell-level anomaly scores: $\bm{S}_{\text{cell}} = [s_{i,j}]$ \\
    %%%
    $\bm{U} \in \mathbb{R}^{m \times d}$ & Uncertainty matrix: $\bm{U} = [u_{i,j}]$ \\
    $\bm{W} \in \mathbb{R}^{m \times d}$ & Confidence weights: $\bm{W} = [w_{i,j}]$ \\
    $\bm{s}_{\text{row}} \in \mathbb{R}^{m}$ & Row-level anomaly scores: $\bm{s}_{\text{row}} = [s_{\text{row},1}, \cdots, s_{\text{row},m}]$  \\
    \bottomrule
\end{tabular}}
\end{table}
\setlength{\textfloatsep}{1.5em}

Let $\bm{X}_{\text{train}} \in \mathbb{R}^{n \times d}$ be a training set with $n$ samples (rows) and $d$ features (columns), and $\bm{X}_{\text{test}} \in \mathbb{R}^{m \times d}$ be a test set with $m$ samples.
We denote the $i$-th sample as $\bm{x}_i = [x_{i,1}, \cdots, x_{i,d}]$, the $j$-th feature as $\bm{x}^j = [x_{1,j}, \cdots, x_{n,j}]$, and the individual cell as $x_{i,j}$.
%%%
The objective of outlier detection is to quantify how anomalous each test sample $\bm{x}_i \in \bm{X}_{\text{test}}$ is, by computing an anomaly score.
%%%
To support both fine-grained analysis and robust detection, we adopt a two-stage scoring approach: first assigning cell-level anomaly scores, then aggregating them into row-level scores.
\begin{definition}[Outlier Detection]
\label{def:outlier-detection}
  Given a test set $\bm{X}_{\text{test}} = [x_{i,j}] \in \mathbb{R}^{m \times d}$, the goal is to compute a cell-level anomaly score $s_{i,j}$ for each feature value $x_{i,j}$, producing a cell-level anomaly matrix:
  \begin{displaymath}
    \bm{S}_{\text{cell}} = [s_{i,j}] \in \mathbb{R}^{m \times d}.
  \end{displaymath}
  %%%
  The anomaly score for row $\bm{x}_i$ is then computed via an aggregation function $g: \mathbb{R}^{d} \to \mathbb{R}$, as follows:
  \begin{displaymath}
    s_{\text{row},i} = g(\bm{s}_{i}) =  g([s_{i,1}, \cdots, s_{i,d}]), 
  \end{displaymath}
  yielding a row-level anomaly score vector:
  \begin{displaymath}
    \bm{s}_{\text{row}} = [s_{\text{row},1}, \cdots, s_{\text{row},m}] \in \mathbb{R}^m.
  \end{displaymath}
  Higher values of $s_{\text{row},i}$ indicate greater likelihoods that $\bm{x}_i$ is an outlier.
\end{definition}

This hierarchical formulation supports both detection accuracy and interpretability: cell-level scores reveal which features contribute to an anomaly, while row-level scores provide a global ranking.
We adopt this structure throughout the paper. 
A summary of frequently used notations is provided in Table~\ref{tab:notation}.

\section{\textsf{RFOD} Framework}
\label{sect:methods}

\begin{figure*}[t]
  \centering
  \includegraphics[width=0.99\textwidth]{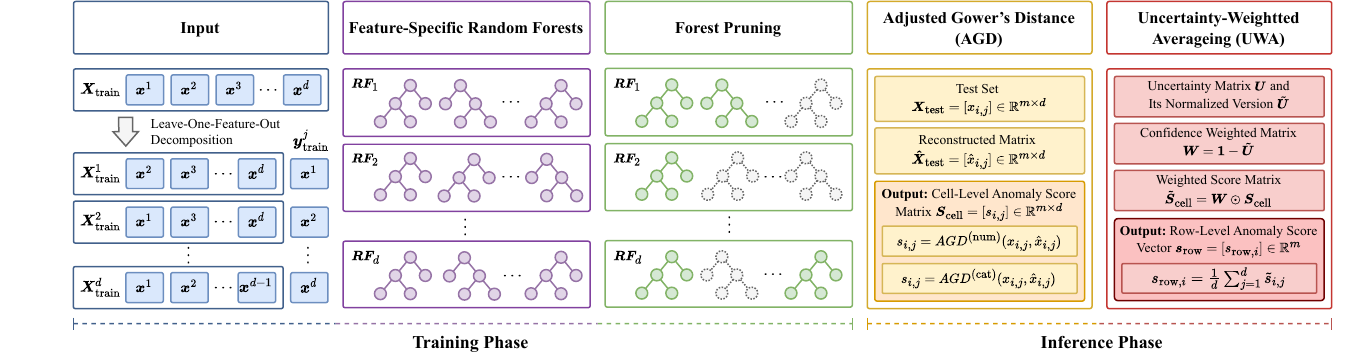}
  \vspace{-1.0em}
  \caption{Overview of \textsf{RFOD}. Given training data $\bm{X}_{\text{train}} \in \mathbb{R}^{n \times d}$, \textsf{RFOD} applies a leave-one-feature-out strategy, training a dedicated forest $\bm{RF}_j$ for each feature $\bm{x}^j$ using the remaining features. Each forest is pruned via out-of-bag (OOB) validation to enhance generalization. At inference, the pruned forests reconstruct $\bm{X}_{\text{test}}$ to form $\hat{\bm{X}}_{\text{test}}$, which is compared against $\bm{X}_{\text{test}}$ using Adjusted Gower's Distance (AGD) to compute cell-level anomaly scores $\bm{S}_{\text{cell}}$. These are aggregated into row-level scores $\bm{s}_{\text{row}}$ via Uncertainty-Weighted Averaging (UWA) for interpretable and robust detection.}
  \label{fig:rfod}
  \vspace{-0.5em}
\end{figure*}

We present \textsf{RFOD}, a robust and interpretable framework for detecting anomalies in tabular data.
%%%
Unlike traditional approaches that model a global joint distribution, often brittle in heterogeneous or high-dimensional settings, \textsf{RFOD} reframes anomaly detection as a \textbf{feature-wise conditional reconstruction} problem. 
It learns how each feature behaves when conditioned on the others, enabling flexible, scalable, and context-aware anomaly detection.

%%%
Given a training set $\bm{X}_{\text{train}} \in \mathbb{R}^{n \times d}$ with $n$ samples and $d$ features, \textsf{RFOD} performs a leave-one-feature-out decomposition.
%%%
For each feature $\bm{x}^j$ ($1 \leq j \leq d$), it builds a dedicated random forest $\bm{RF}_j$ that treats $\bm{x}^j$ as the target and uses the remaining $(d-1)$ features as inputs.
This results in $d$ independent forests $\{\bm{RF}_1, \cdots, \bm{RF}_d\}$, each capturing conditional patterns specific to a feature. 
%%%
These forests are trained in parallel for efficiency.

%%%
During inference, each test sample $\bm{x}_i = [x_{i,1}, \cdots, x_{i,d}]$ from $\bm{X}_{\text{test}} \in \mathbb{R}^{m \times d}$ is evaluated by the $d$ trained forests. 
For each feature $x_{i,j}$, the corresponding forest $\bm{RF}_j$ predicts a value $\hat{x}_{i,j}$ based on the remaining features. 
%%%
In normal cases, $\hat{x}_{i,j}$ should closely match $x_{i,j}$, reflecting consistency with learned patterns.
In contrast, anomalies often exhibit larger deviations between $x_{i,j}$ and $\hat{x}_{i,j}$, resulting in higher residuals and anomaly scores.
%%%
Notably, even when $\hat{x}_{i,j}$ appears accurate, inconsistencies in decision paths, such as diverging leaf nodes across trees, may signal contextual anomalies.
This enables \textsf{RFOD} to capture overt and subtle deviations from normality.

As illustrated in Figure~\ref{fig:rfod}, \textsf{RFOD} comprises four key modules:
\begin{enumerate}[nolistsep,leftmargin=24pt]
  \item \textbf{Feature-Specific Random Forests} (\sec{\ref{sect:methods:module1}}): 
  Train a separate forest for each feature, using the others as input, to learn conditional distributions.
  
  \item \textbf{Forest Pruning} (\sec{\ref{sect:methods:module2}}): 
  Retain only the most informative trees using out-of-bag (OOB) validation, controlled by a retaining ratio $\beta$.
  
  \item \textbf{Adjusted Gower's Distance (AGD)} (\sec{\ref{sect:methods:module3}}): 
  Compare actual and predicted values using AGD to produce a cell-level anomaly score matrix.
  
  \item \textbf{Uncertainty-Weighted Averaging (UWA)} (\sec{\ref{sect:methods:module4}}): 
  Use tree disagreement to estimate uncertainty, which weights cell scores when aggregating to row-level scores.
\end{enumerate}

This architecture enables robust, interpretable, and scalable anomaly detection across diverse tabular domains.

\subsection{Feature-Specific Random Forests}
\label{sect:methods:module1}

%%%
To capture complex inter-feature dependencies, \textsf{RFOD} employs a leave-one-feature-out decomposition strategy. 
Instead of fitting a single global model across all features, it constructs a set of feature-specific random forests, each dedicated to learning the conditional behavior of one feature given the rest.

\paragraph{Leave-One-Feature-Out Decomposition}
%%%
For each feature $\bm{x}^j$ ($1 \leq j \leq d$) in the training set $\bm{X}_{\text{train}} \in \mathbb{R}^{n \times d}$, \textsf{RFOD} treats $\bm{x}^j$ as the target and uses the remaining $(d - 1)$ features as predictors.
This results in $d$ overlapping sub-training sets of the form $(\bm{X}_{\text{train}}^j, \bm{y}_{\text{train}}^j)$, where $\bm{X}_{\text{train}}^j = \bm{X}_{\text{train}} \setminus \{\bm{x}^j\}$ and $\bm{y}_{\text{train}}^j = \bm{x}^j$.
%%%
Each sub-training set is used to train a dedicated Random Forest $\bm{RF}_j$ that approximates the conditional distribution of the target feature:
\begin{equation}
\label{eqn:rf-model}
\bm{RF}_j: \bm{X}^j_{\text{train}} \in \mathbb{R}^{n \times (d-1)} \rightarrow \bm{y}_{\text{train}}^j \in \mathbb{R}^{n}.
\end{equation}

%%%
The model type is selected based on the target feature's data type: Random Forest classifiers are used for categorical features and regressors for numerical ones (see Lines \ref{algo:cate}--\ref{algo:reg} in Algorithm~\ref{algo:rfod}).
%%%
Each $\bm{RF}_j$ learns the typical behavior of its feature under normal conditions, conditioned on the context of other features.
Together, the $d$ forests encode the feature-wise conditional distributions that define the structure of normal data.

\paragraph{Remarks}
This design offers three key advantages:
(1) By modeling each feature conditionally, \textsf{RFOD} captures both global and local inter-feature dependencies, enabling the detection of subtle context-specific anomalies.
(2) Avoiding full joint modeling mitigates the curse of dimensionality, which often hinders anomaly detection in high-dimensional spaces~\cite{zhao2021suod}. 
(3) The independent training of each forest allows for straightforward parallelization, significantly improving efficiency--a benefit we empirically validate in \sec{\ref{sect:expt:scalability}}.

\subsection{Forest Pruning}
\label{sect:methods:module2}

\paragraph{Motivation}
Although Random Forests are known for their robustness, they often include a large number of decision trees, many of which contribute little to predictive accuracy. 
These low-quality or redundant trees not only increase memory and inference costs but can also introduce noise that undermines generalization.
To mitigate these issues, \textsf{RFOD} integrates a pruning module that selectively retains the most informative trees from each feature-specific forest, thereby enhancing both efficiency and model robustness.

\begin{figure*}[t]
\centering
\subfigure[Small $\beta$ (only a few trees are informative).]{%
  \label{fig:agd_num_example:small}%
  \includegraphics[width=0.33\textwidth]{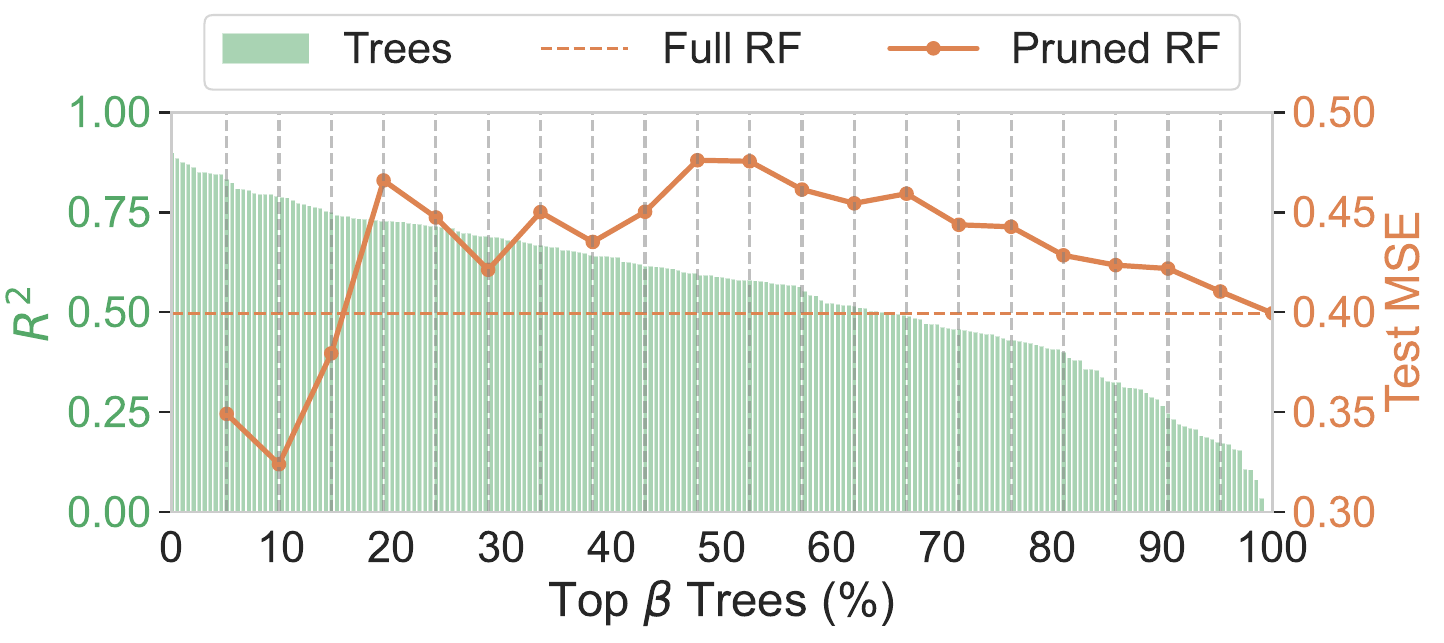}}%
\subfigure[Moderate $\beta$ (roughly half of the trees are useful).]{%
  \label{fig:agd_num_example:medium}%
  \includegraphics[width=0.33\textwidth]{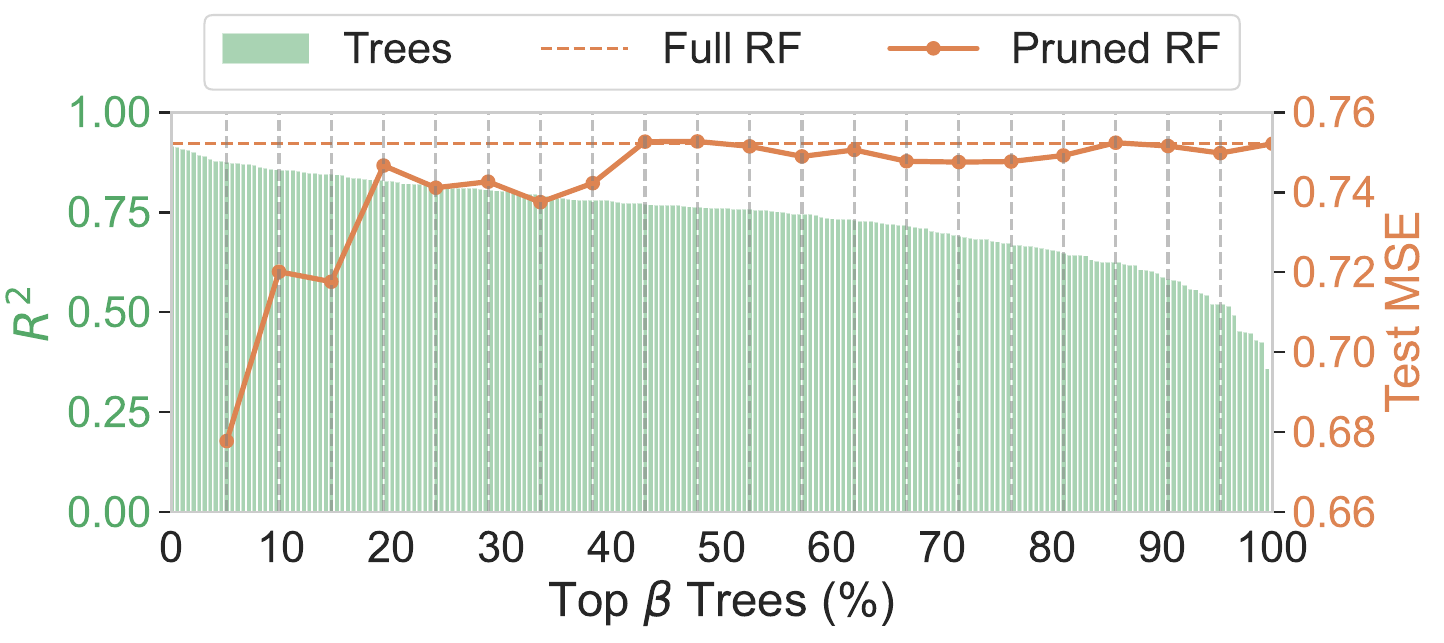}}%
\subfigure[Large $\beta$ (tree quality is uniform).]{%
  \label{fig:agd_num_example:large}%
  \includegraphics[width=0.33\textwidth]{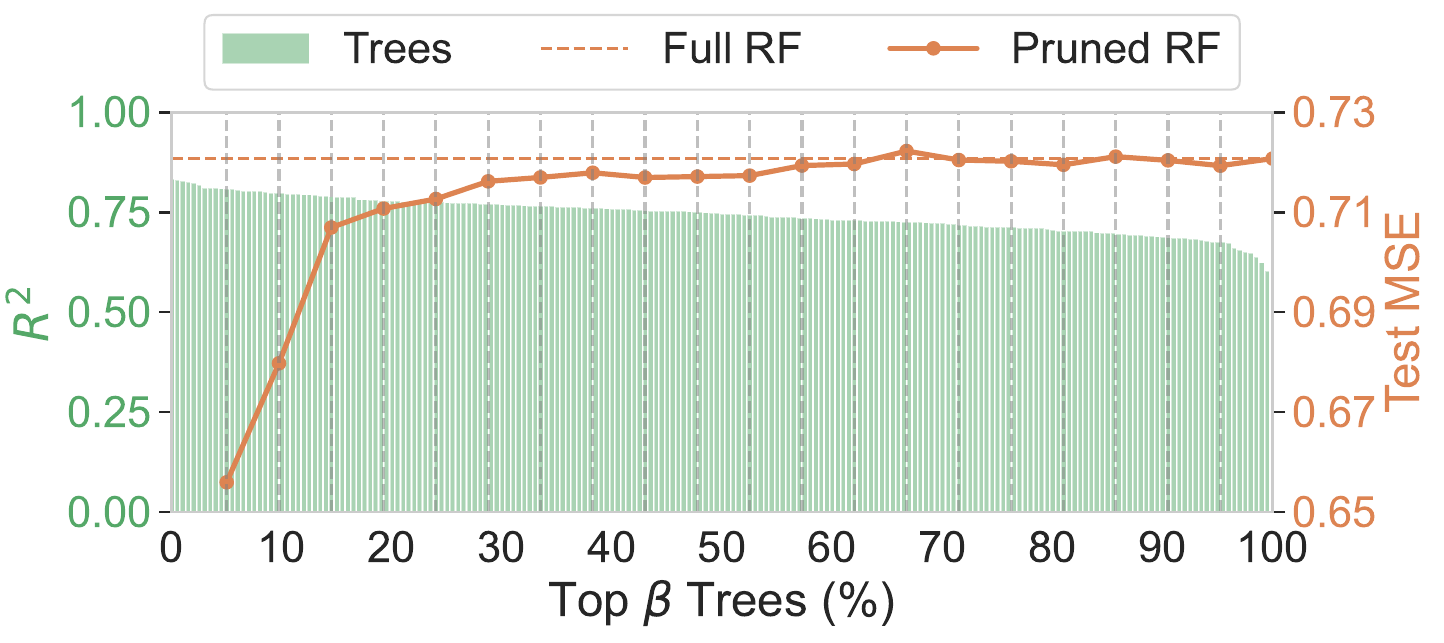}}%
\vspace{-1.0em}%
\caption{Illustration of forest pruning with varying retaining ratio $\beta$.}%
\label{fig:pruning}%
\vspace{-0.75em}
\end{figure*}

\paragraph{Pruning Strategy}
Given a feature-specific forest $\bm{RF} = \{T_1, \cdots, T_t\}$ with $t$ decision trees, \textsf{RFOD} applies a pruning strategy that leverages out-of-bag (OOB) validation--a built-in mechanism of Random Forests.
%%%
Each tree $T_i$ ($1 \leq i \leq t$) is evaluated on its OOB samples using either AUC-ROC (for classification tasks) or the coefficient of determination $R^2$ (for regression tasks).
Formally, the OOB score $\phi(T_i)$ is computed as:
\begin{displaymath}
  \phi(T_i) = \begin{cases}
    \text{AUC-ROC}(T_i), & \text{for classification} \\
    R^2(T_i). & \text{for regression}
  \end{cases}
\end{displaymath}

The trees are sorted in descending order of $\phi(T_i)$, yielding an index permutation $U(1), \cdots, U(t)$ such that $\phi(T_{U(1)}) \geq \cdots \geq \phi(T_{U(t)})$. 
%%%
The top $\lfloor \beta \cdot t \rfloor$ trees are retained:
\begin{displaymath}
  \text{Prune}(\bm{RF}) = \left\{ T_{U(i)} \mid 1 \leq i \leq \floor{\beta \cdot t} \right\},
\end{displaymath}
where $\beta \in (0,1]$ is the retaining ratio (see Line~\ref{algo:prune} in Algorithm~\ref{algo:rfod}).

This pruning procedure compresses the model and reduces inference overhead without sacrificing--and often improving--detection performance. 
As shown in the ablation study (\sec{\ref{sect:expt:ablation}}), carefully pruned forests frequently outperform their unpruned counterparts, particularly in high-dimensional settings.

\begin{example}
  Figure~\ref{fig:pruning} illustrates the effect of varying the pruning ratio $\beta \in [5\%, 100\%]$ (in 5\% increments) on model performance across three representative numerical prediction tasks.
  For each task, decision trees are ranked by their out-of-bag (OOB) $R^2$ scores, and subsets comprising the top-ranked trees are retained.
  In each subfigure, green bars (left $y$-axis) indicate the $R^2$ of individual trees. The orange solid curve (right $y$-axis) shows the test MSE of the pruned forest as $\beta$ increases, while the dashed orange line denotes the baseline test MSE of the full (unpruned) forest.
  \begin{itemize}[nolistsep,leftmargin=20pt]
    \item \textbf{Small $\beta$} (Figure~\ref{fig:agd_num_example:small}): 
    When only a few trees are informative, retaining the top 20\% reduces test MSE by about 16.6\%. Including more low-quality trees leads to performance degradation.
    
    \item \textbf{Moderate $\beta$} (Figure~\ref{fig:agd_num_example:medium}): 
    When roughly half of the trees are useful, performance stabilizes around $\beta=0.5$, achieving a favorable trade-off between model size and accuracy.
    
    \item \textbf{Large $\beta$} (Figure \ref{fig:agd_num_example:large}): 
    When tree quality is relatively uniform, pruning yields marginal gains. Nevertheless, using a moderately high $\beta$ (e.g., 0.8) still reduces model size without sacrificing accuracy.
  \end{itemize}
  
  %%%
  In summary, this adaptive pruning strategy aligns with the underlying distribution of tree quality--pruning aggressively when the ensemble is noisy and conservatively when tree quality is consistent. 
  This flexibility enables \textsf{RFOD} to maintain robust performance while improving efficiency.
  \hfill $\triangle$ \par
\end{example}

\subsection{Adjusted Gower's Distance}
\label{sect:methods:module3}

After forest pruning, the model predicts each entry of the test set $\bm{X}_{\text{test}} = [x_{i,j}] \in \mathbb{R}^{m \times d}$, producing the reconstruction matrix $\bm{\hat{X}}_{\text{test}} = [\hat{x}_{i,j}] \in \mathbb{R}^{m \times d}$. 
Each predicted value $\hat{x}_{i,j}$ estimates the actual cell $x_{i,j}$ based on conditional patterns learned from normal data.
%%%
To identify anomalies, \textsf{RFOD} computes a fine-grained anomaly score matrix $\bm{S}_{\text{cell}} = [s_{i,j}] \in \mathbb{R}^{m \times d}$ by comparing each actual value to its prediction.
Each score $s_{i,j}$ reflects the degree of deviation for a single feature, offering interpretability at the cell level.

\paragraph{Gower's Distance (GD) and Its Limitations} 
%%%
A key challenge is choosing a distance metric suitable for mixed-type tabular data.
Traditional metrics like Euclidean or cosine distance assume homogeneous feature types and overlook prediction uncertainty.
Gower's Distance (GD)~\cite{gower1971general} is more appropriate, as it handles both numerical and categorical features:
\begin{itemize}[nolistsep,leftmargin=20pt]
  \item For \textbf{numerical features}, it employs min-max normalization:
  \begin{displaymath}
    {GD}^{(\text{num})}(x_{i,j}, \hat{x}_{i,j}) = \frac{|x_{i,j} - \hat{x}_{i,j}|}{\max(\bm{x}^j) - \min(\bm{x}^j)},
  \end{displaymath}
  while the IQR-based variant \cite{iqrgower} improves robustness by replacing the denominator with the interquartile range:
  \begin{displaymath}
    {GD}_{\text{IQR}}^{(\text{num})}(x_{i,j}, \hat{x}_{i,j}) = \frac{|x_{i,j} - \hat{x}_{i,j}|}{Q_{0.75}(\bm{x}^j) - Q_{0.25}(\bm{x}^j)},
  \end{displaymath}
  where $Q_{0.25}$ and $Q_{0.75}$ are the 25- and 75-quantiles of feature $\bm{x}^j$, respectively.

  \item For \textbf{categorical features}, GD is defined by binary matching:
  \begin{displaymath}
    {GD}^{(\text{cat})}(x_{i,j}, \hat{x}_{i,j}) = \begin{cases}
      0, & \text{if}~x_{i,j} = \hat{x}_{i,j} \\
      1. & \text{otherwise}
    \end{cases}
  \end{displaymath}
\end{itemize}

The final distance is the average of these per-feature scores.

Despite its widespread use, GD suffers from two key limitations.
First, for numerical features, standard scaling methods such as min-max and IQR are sensitive to skewed or heavy-tailed distributions, which can distort the computed dissimilarities.
Second, for categorical features, GD relies on binary matching, disregarding model confidence and thereby failing to differentiate between confident and uncertain predictions.

\paragraph{Adjusted Gower's Distance (AGD)}
To address these issues, we propose Adjusted Gower's Distance (AGD), a metric tailored for anomaly detection in tabular data. It incorporates:
\begin{itemize}[nolistsep,leftmargin=20pt]
  \item \textbf{$\alpha$-Quantile Scaling:} 
  A generalization of IQR scaling for numerical features that adapts to distributional asymmetry.
  
  \item \textbf{Confidence-Aware Scoring:} 
  A soft matching scheme for categorical features based on class probability.
\end{itemize}

This formulation yields robust and interpretable scores even under skewed distributions and prediction uncertainty.
Definitions for numerical and categorical features are given below.
\begin{definition}[AGD for Numerical Features]
\label{def:agd_numerical}
  For numerical features, AGD is defined via adaptive $\alpha$-quantile scaling:
  \begin{equation}
    \text{AGD}^{(\text{num})}(x_{i,j}, \hat{x}_{i,j}) = \frac{|x_{i,j} - \hat{x}_{i,j}|}{Q_{1-\alpha}(\bm{x}^j) - Q_{\alpha}(\bm{x}^j)},
  \end{equation}
  where $\alpha \in (0, 0.5)$ controls sensitivity to distribution tails; $Q_{\alpha}$ and $Q_{1-\alpha}$ are the $\alpha$- and $(1-\alpha)$-quantiles of feature $\bm{x}^j$, respectively. 
\end{definition}

\begin{figure*}[t]
\centering
\subfigure[Skewed distribution of a numerical feature.]{%
  \label{fig:agd_num_example:1}%
  \includegraphics[width=0.399213088\textwidth]{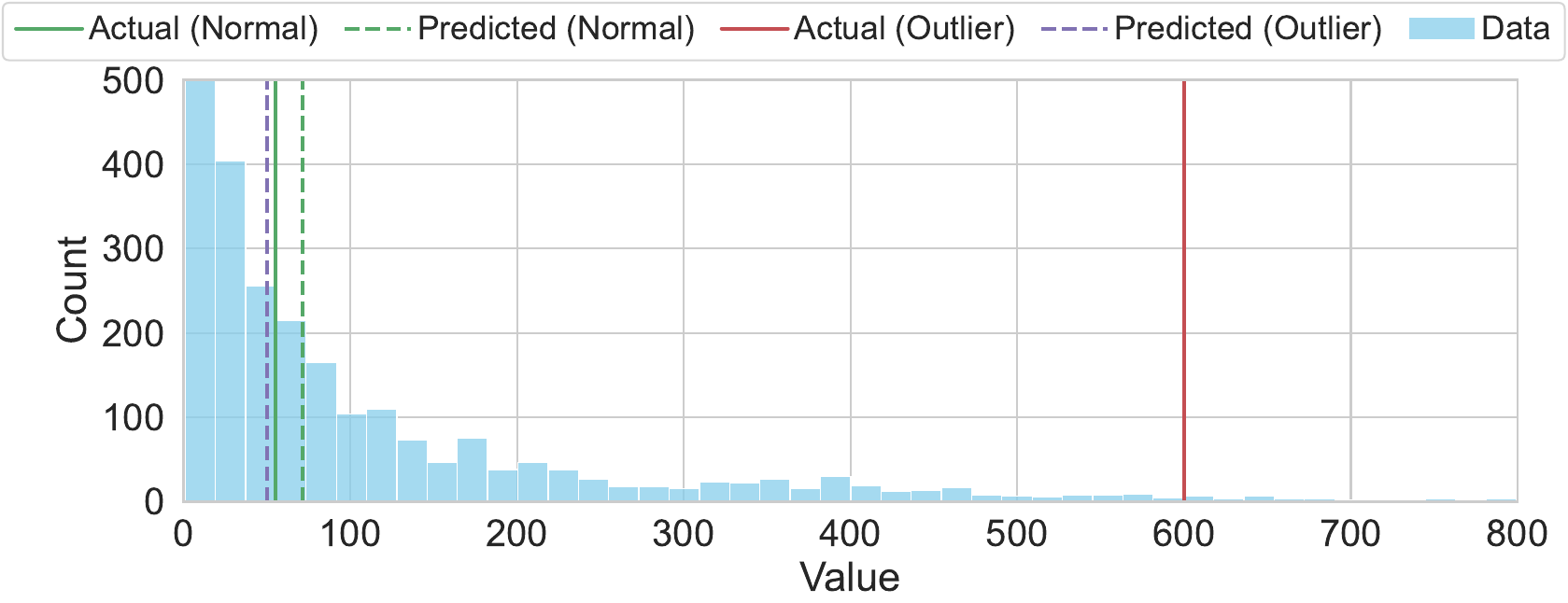}}%
\subfigure[Comparison of different distance measures.]{% 
  \label{fig:agd_num_example:2}%
  \includegraphics[width=0.263848538\textwidth]{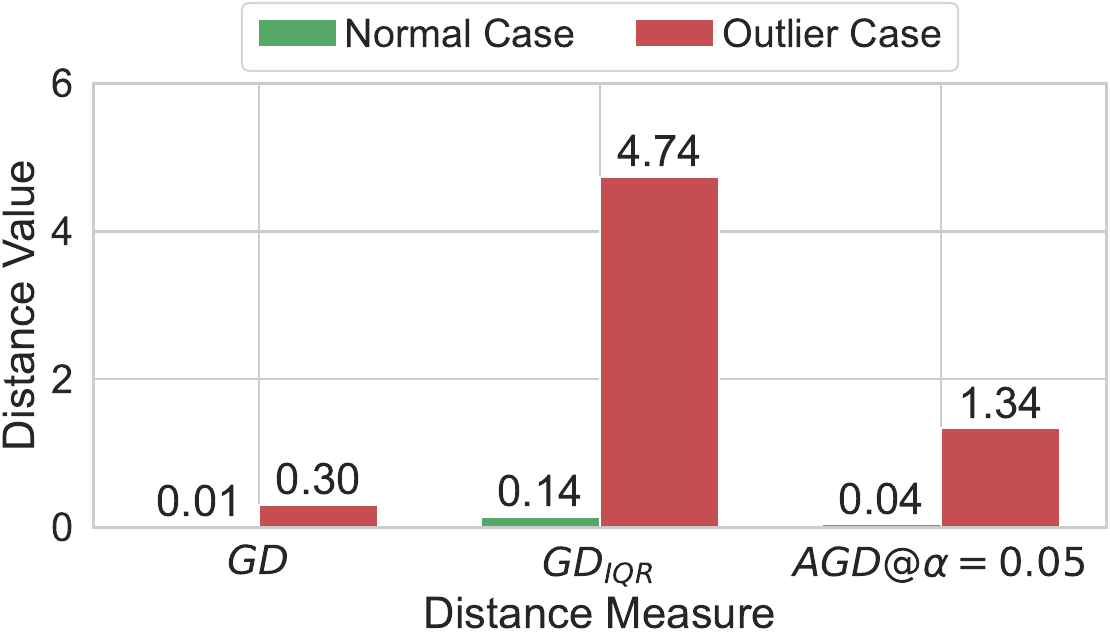}}%
\subfigure[Sensitivity of AGD to different $\alpha$ values.]{%
  \label{fig:agd_num_example:3}%
  \includegraphics[width=0.327037384\textwidth]{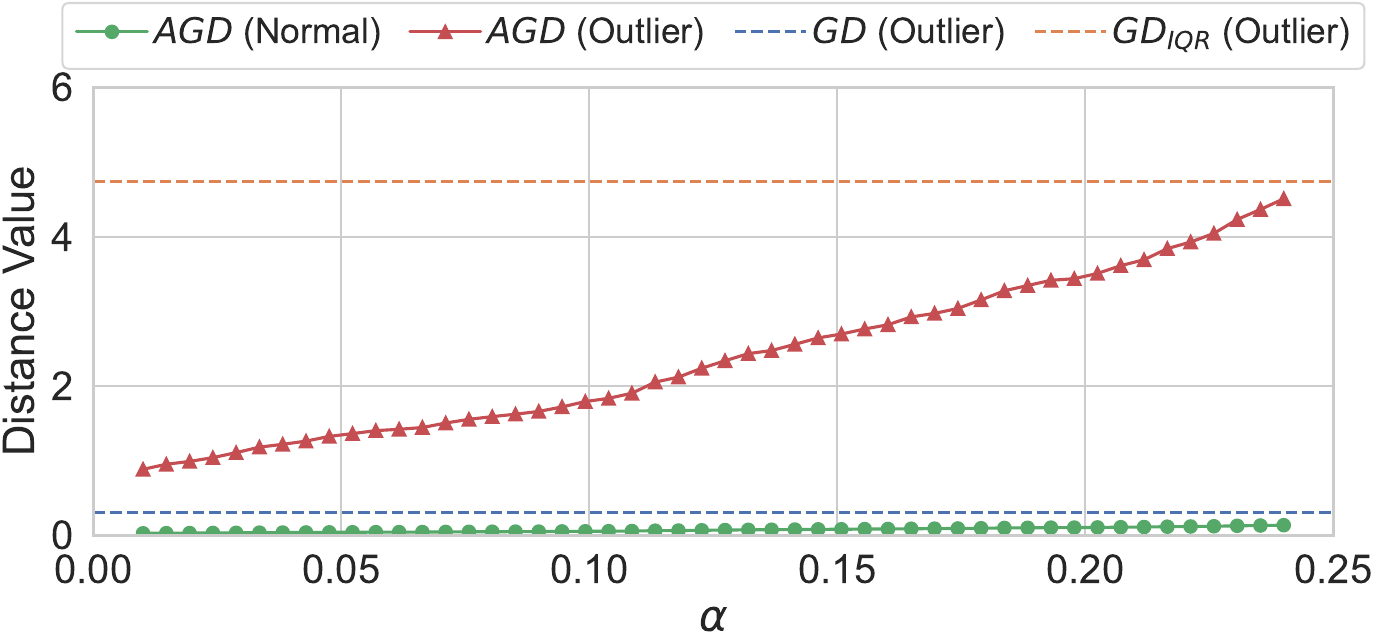}}%
\vspace{-1.25em}%
\caption{Challenges of existing Gower's Distance (GD) and the proposed AGD for numerical features.}%
\label{fig:agd_num_example}%
\vspace{-0.25em}
\end{figure*}

\vspace{-0.5em}
\begin{example}[AGD for Numerical Features]
  Figure~\ref{fig:agd_num_example} illustrates the behavior of Adjusted Gower's Distance (AGD) on a skewed numerical feature from the Pageblocks dataset \cite{page_blocks_classification_78}.
  %%%
  In Figure~\ref{fig:agd_num_example:1}, we show the feature's value distribution, highlighting actual and predicted values for both a normal and an outlier instance. 
  %%%
  For the normal case, the predicted value (green dashed line) closely aligns with the actual value (green solid line), indicating consistency with learned patterns. In contrast, the outlier's prediction (purple dashed line) substantially deviates from its actual value (red solid line), reflecting an anomalous behavior.

  Figure \ref{fig:agd_num_example:2} compares the distance values computed by traditional GD, IQR-based GD, and AGD (with $\alpha=0.05$). 
  Traditional GD underestimates the anomaly score (0.30), while the IQR-based variant overshoots with an inflated value (4.74). 
  AGD provides a more balanced assessment, assigning a low score (0.04) to the normal instance and a higher, yet reasonable, score (1.34) to the outlier.
  
  Lastly, Figure~\ref{fig:agd_num_example:3} visualizes how the choice of $\alpha$ influences AGD's sensitivity.
  Smaller $\alpha$ values yield behavior similar to traditional GD, while larger values resemble IQR-based scoring. A comprehensive sensitivity analysis is provided in \sec{\ref{sect:expt:hyperparam}}.
  \hfill $\triangle$ \par
\end{example}

\begin{definition}[AGD for Categorical Features]
\label{def:agd_categorical}
For categorical features, AGD transforms binary matching into prediction confidence:
\begin{equation}
\text{AGD}^{(\text{cat})}(\bm{x}_{i,j}, \hat{x}_{i,j}) = 1 - p_{x_{i,j}},
\end{equation}
where $p_{x_{i,j}}$ is the predicted probability of the true category $x_{i,j}$.
This formulation reflects not only correctness but also model certainty.
\end{definition}

\begin{example}[AGD for Categorical Features]
  Table~\ref{tab:categ_example} illustrates how AGD captures prediction confidence for categorical features.
  %%%
  For correct predictions (Rows 1, 2, and 4), GD assigns a fixed score of 0, ignoring confidence levels. In contrast, AGD reflects varying certainty by assigning lower scores to more confident predictions.
  %%%
  For misclassifications (Rows 3, 5, and 6), GD gives a score of 1 regardless of prediction quality, whereas AGD varies the score based on the confidence of the incorrect prediction.
  \hfill $\triangle$ \par  
\end{example}

\begin{table}[ht]
\centering
\small
\setlength\tabcolsep{3pt}
\renewcommand{\arraystretch}{1.1}
\caption{Illustrative examples of probability-based adjustments in AGD for categorical features, highlighting how model confidence influences dissimilarity scoring.}
\label{tab:categ_example}
\vspace{-1.0em}
\resizebox{\columnwidth}{!}{%
\begin{tabular}{ccccc}
  \toprule
  \rowcolor[HTML]{FFF2CC}
  $x_{i,j}$ & $\hat{x}_{i,j}$ & Predicted Probability & ${GD}^{(\text{cat})}(x_{i,j}, \hat{x}_{i,j})$ & ${AGD}^{(\text{cat})}(x_{i,j}, \hat{x}_{i,j})$ \\
  \midrule
  $A$ & $A$ & $p_A=0.8$, $p_B=0.2$ & 0 & 1 -- 0.8 = 0.2 \\
  $A$ & $A$ & $p_A=0.6$, $p_B=0.4$ & 0 & 1 -- 0.6 = 0.4 \\
  $A$ & $B$ & $p_A=0.3$, $p_B=0.7$ & 1 & 1 -- 0.3 = 0.7 \\
  \midrule
  \rowcolor[HTML]{DDEBFF} $A$ & $A$ & $p_A=0.5$, $p_B=0.3$, $p_C=0.2$ & 0 & 1 -- 0.5 = 0.5 \\
  \rowcolor[HTML]{DDEBFF} $A$ & $C$ & $p_A=0.1$, $p_B=0.2$, $p_C=0.7$ & 1 & 1 -- 0.1 = 0.9 \\
  \rowcolor[HTML]{DDEBFF} $A$ & $B$ & $p_A=0.2$, $p_B=0.7$, $p_C=0.1$ & 1 & 1 -- 0.2 = 0.8 \\
  \bottomrule
\end{tabular}}
\end{table}

\paragraph{Cell-Level Scoring}
Finally, to compute cell-level anomaly scores, \textsf{RFOD} applies AGD between the original test matrix $\bm{X}_{\text{test}} = [x_{i,j}]$ and its reconstructed counterpart $\bm{\hat{X}}_{\text{test}} = [\hat{x}_{i,j}]$.
This yields the complete cell-level anomaly score matrix $\bm{S}_{\text{cell}} = [s_{i,j}] \in \mathbb{R}^{m \times d}$, where each score is defined as:
\begin{displaymath}
  s_{i,j} = \begin{cases}
    \text{AGD}^{(\text{num})}(x_{i,j}, \hat{x}_{i,j}), & \text{if $x_{i,j}$ is numerical} \\
    \text{AGD}^{(\text{cat})}(x_{i,j}, \hat{x}_{i,j}). & \text{if $x_{i,j}$ is categorical}
  \end{cases}
\end{displaymath}

High anomaly scores indicate feature values that deviate significantly from their expected behavior, allowing \textsf{RFOD} to pinpoint not only anomalous rows but also the exact cells responsible. 
This fine-grained interpretability is critical for tasks like root cause analysis and transparent anomaly attribution~\cite{li2022ecod, sun2025unifying, cad}.

\subsection{Uncertainty-Weighted Averaging}
\label{sect:methods:module4}

\paragraph{Unweighted Averaging}
While cell-level anomaly scores provide fine-grained interpretability, many real-world applications require a single anomaly score per row \cite{li2022ecod, dif, chandola2009adsurvey}. 
%%%
A simple approach is to average the cell-level scores across each row without weighting.
%%%
Yet, this naive method assumes equal reliability across all features, which is often unrealistic in heterogeneous tabular data.
As a result, noisy or uncertain predictions may distort the row-level score, either inflating noise or masking true anomalies.

\paragraph{Uncertainty-Weighted Averaging (UWA)}
To address this, we propose Uncertainty-Weighted Averaging (UWA), an adaptive aggregation scheme that adjusts each cell's influence based on the model's predictive confidence.
%%%
Leveraging the ensemble structure of Random Forests, we quantify uncertainty as the standard deviation (STD) of predictions across decision trees: higher STDs indicate greater uncertainty and result in lower weights, while lower STDs imply confidence and are upweighted accordingly.

As described in Algorithm~\ref{algo:rfod} (Lines \ref{algo:row_std}--\ref{algo:row_score}), we first compute an uncertainty matrix $\bm{U} = [u_{i,j}] \in \mathbb{R}^{m \times d}$, where each $u_{i,j}$ denotes the STD of predictions for $\hat{x}_{i,j}$.
Each row $\bm{u}_i$ is then normalized to yield a row-wise uncertainty profile:
\begin{displaymath}
  \tilde{\bm{u}}_i = \frac{\bm{u}_i}{\sum_{j=1}^d u_{i,j}},~\forall i \in \{1,\cdots,m\}.
\end{displaymath} 
%%%
The resulting matrix $\tilde{\bm{U}} = [\tilde{\bm{u}}_1, \cdots, \tilde{\bm{u}}_m]$ is subtracted from an all-one matrix to produce the confidence weight matrix $\bm{W} \in \mathbb{R}^{m \times d}$:
\begin{displaymath}
  \bm{W} = \bm{1}^{m \times d} - \tilde{\bm{U}}.
\end{displaymath}
Lower uncertainty leads to higher confidence weights.
%%%
Next, given $\bm{S}_{\text{cell}} = [s_{i,j}] \in \mathbb{R}^{m \times d}$, we compute a weighted score matrix $\tilde{\bm{S}}_{\text{cell}} = [\tilde{s}_{i,j}] \in \mathbb{R}^{m \times d}$ via element-wise multiplication:
\begin{displaymath}
  \tilde{\bm{S}}_{\text{cell}} = \bm{W} \odot \bm{S}_{\text{cell}}.
\end{displaymath}
%%%
The final row-level anomaly score vector $\bm{s}_{\text{row}} = [s_{\text{row},1}, \cdots, s_{\text{row},m}]$ is computed by averaging the weighted scores across each row:
\begin{displaymath}
  s_{\text{row},i} = \frac{1}{d} \sum_{j=1}^d \tilde{s}_{i,j},~\forall i \in \{1, \cdots, m\}.
\end{displaymath}

\paragraph{Remarks}
Compared to Unweighted Averaging, UWA offers three advantages. 
(1) It incorporates model confidence into the scoring process, allowing for more informative and introspective anomaly analysis. 
(2) It naturally downweights unreliable predictions, thereby enhancing robustness against noise and uncertainty.
(3) It improves detection performance in scenarios where anomalies affect only a subset of features by emphasizing reliable signals.
We validate these benefits through ablation studies in \sec{\ref{sect:expt:ablation}}.

\subsection{Overall Algorithm}
\label{sect:methods:algo}

The complete workflow of \textsf{RFOD} is outlined in Algorithm~\ref{algo:rfod}. 
%%%
Given a training set $\bm{X}_{\text{train}}$ (assumed to contain only normal samples), a test set $\bm{X}_{\text{test}}$ (which may contain anomalies), and two hyperparameters, $\alpha$ for AGD sensitivity and $\beta$ for forest pruning, the algorithm outputs both a cell-level anomaly score matrix $\bm{S}_{\text{cell}}$ and a row-level anomaly score vector $\bm{s}_{\text{row}}$.

\paragraph{Training Phase}
For each feature $\bm{x}^j \in \bm{X}_{\text{train}}$ ($1 \leq j \leq d$), \textsf{RFOD} performs a leave-one-feature-out decomposition by treating $\bm{x}^j$ as the prediction target $\bm{y}_{\text{train}}^j$ and trains a feature-specific Random Forest $\bm{RF}_j$ using the rest features $\bm{X}_{\text{train}}^j = \bm{X}_{\text{train}} \setminus \{\bm{x}^j\}$ as input (Line \ref{algo:feature_decomposition}).
The model type, i.e., classifier or regressor, is selected based on whether the target feature is categorical or numerical (Lines~\ref{algo:cate}--\ref{algo:reg}).
%%%
To improve generalization and reduce inference overhead, each forest $\bm{RF}_j$ is then pruned by retaining only the top $\lfloor \beta \cdot t \rfloor$ trees, selected via OOB validation (Line~\ref{algo:prune}).

\paragraph{Inference Phase}
During inference, each pruned forest $\bm{RF}_j$ predicts its corresponding feature $x_{i,j}$ in the test set $\bm{X}_{\text{test}}$, forming the reconstruction matrix $\hat{\bm{X}}_{\text{test}}$ and an associated uncertainty matrix $\bm{U}$, where each $u_{i,j}$ captures the standard deviation across tree predictions (Lines \ref{algo:prediction_start}--\ref{algo:prediction_aggregation}).
%%%
Using the original and reconstructed values, \textsf{RFOD} computes the cell-level anomaly scores $\bm{S}_{\text{cell}}$ via AGD, applied separately to numerical and categorical features as defined in Definitions~\ref{def:agd_numerical} and~\ref{def:agd_categorical} (Line \ref{algo:cell_score}).
%%%
To obtain row-level anomaly scores, the uncertainty matrix $\bm{U}$ is row-normalized to produce confidence weights $\bm{W}$ (Lines~\ref{algo:row_std}--\ref{algo:row_confidence}), which are then used to reweight $\bm{S}_{\text{cell}}$ and compute the final scores $\bm{s}_{\text{row}}$ (Lines \ref{algo:adj_cell}--\ref{algo:row_score}).

\begin{algorithm}[t]
\caption{\textsf{RFOD}}
\label{algo:rfod}
\KwIn{Training set $\bm{X}_{\text{train}} \in \mathbb{R}^{n \times d}$; 
  Test set $\bm{X}_{\text{test}} \in \mathbb{R}^{m \times d}$; 
  Quantile parameter $\alpha$ for Adjusted Gower's Distance (AGD); 
  Retaining ratio $\beta$ for forest pruning\;
}
\KwOut{Cell-level anomaly score matrix $\bm{S}_{\text{cell}} \in \mathbb{R}^{m \times d}$;
  Row-level anomaly score vector $\bm{s}_{\text{row}} \in \mathbb{R}^m$\;
}
%%% initialization
$\bm{\hat{X}}_{\text{test}} \gets \mathbf{0}^{m \times d}$\tcp*{Reconstructed test set} 
$\bm{U} \gets \mathbf{0}^{m \times d}$\tcp*{Uncertainty matrix}
%%% Train feature-specific random forests and make predictions
\For{$j=1$ \KwTo $d$}{
  %%% feature-specific training
  $\bm{X}_{\text{train}}^j \gets \bm{X}_{\text{train}} \setminus \{ \bm{x}^j \}$; $\bm{y}_{\text{train}}^j \gets \bm{x}^j$\; \label{algo:feature_decomposition}
  \eIf{$\bm{y}_{\text{train}}^j$ is categorical}{ \label{algo:cate}
    Train classifier $\bm{RF}_j$ on $(\bm{X}_{\text{train}}^j, \bm{y}_{\text{train}}^j)$\;
  }{
    Train regressor $\bm{RF}_j$ on $(\bm{X}_{\text{train}}^j, \bm{y}_{\text{train}}^j)$\; \label{algo:reg}
  }
  %%% forest pruning
  Prune $\bm{RF}_j$ to retain top $\lfloor \beta \cdot t \rfloor$ trees via OOB validation\; \label{algo:prune}
  %%% inference
  $\{\hat{\bm{x}}^j, \bm{u}^j\} \gets \bm{RF}_j(\bm{X}_{\text{test}}^j)$, where $\bm{X}_{\text{test}}^j \gets \bm{X}_{\text{test}} \setminus \{\bm{x}^j\}$\; \label{algo:prediction_start}
  $\bm{\hat{X}}_{\text{test}}[:, j] \gets \hat{\bm{x}}^j$; $\bm{U}[:, j] \gets \bm{u}^j$\; \label{algo:prediction_aggregation}
}
%%% cell-level
$\bm{S}_{\text{cell}} \gets \text{AGD}(\bm{X}_{\text{test}}, \hat{\bm{X}}_{\text{test}}, \alpha)$\tcp*{Defs~\ref{def:agd_numerical} and \ref{def:agd_categorical}}  \label{algo:cell_score}
%%% row-level
Normalize each row $\bm{u}_i \in \bm{U}$:~$\tilde{\bm{u}}_i \gets \tfrac{\bm{u}_i}{\sum_{j=1}^d u_{i,j}}$,~$\forall i \in \{1, \ldots, m\}$\; \label{algo:row_std}
Compute confidence weights: $\bm{W} \gets \bm{1}^{m \times d} - \tilde{\bm{U}}$\; \label{algo:row_confidence}
$\tilde{\bm{S}}_{\text{cell}} \gets \bm{W} \odot \bm{S}_{\text{cell}}$\tcp*{Element-wise weighting} \label{algo:adj_cell}
$s_{\text{row},i} \gets \frac{1}{d}\sum_{j=1}^d \tilde{s}_{i,j}$,~$\forall i \in \{1, \ldots, m\}$\; \label{algo:row_score}
\Return $\bm{S}_{\text{cell}}$, $\bm{s}_{\text{row}} \gets [s_{\text{row},1}, \cdots, s_{\text{row},m}]$\;
\end{algorithm}

% $\bm{\tilde{U}} = [\tilde{\bm{u}}_1, \cdots, \tilde{\bm{u}}_m]$,

\paragraph{Time Complexity Analysis}
Let $t$ denote the number of decision trees per forest, $n$ the number of training samples, and $d$ the number of features.
%%%
Each tree has an expected depth of $O(\log n)$ under balanced splits. Training a single random forest for one feature thus incurs a cost of $O(tnd \log n)$. 
Since \textsf{RFOD} constructs one forest per feature, the overall training complexity is $O(tnd^2 \log n)$.

During inference, each of the $m$ test samples is evaluated across all $d$ forests, each with $t$ trees. As each tree traversal costs $O(\log m)$, the total inference complexity is $O(tmd \log m)$.

\paragraph{Distributed Learning}
\textsf{RFOD} is inherently well-suited for distributed computation. The leave-one-feature-out design ensures that each of the $d$ forests can be trained independently, and the $t$ trees within each forest can be trained in parallel.
%%%
With $c$ compute units available, the $td$ trees can be evenly partitioned, reducing the per-unit training cost to $O\left(\frac{tnd^2 \log n}{c}\right)$.

Inference is even more amenable to fine-grained parallelism. 
Each prediction of a feature value in a test instance constitutes a lightweight, independent task, yielding a total of $mtd$ atomic tasks, each with a computational complexity of $O(\log m)$.
%%%
We empirically validate these efficiency gains in \sec{\ref{sect:expt:efficiency}}.

\paragraph{Summary of Strengths}
\textsf{RFOD} brings a range of advantages that make it well-suited for real-world anomaly detection tasks:
\begin{itemize}[nolistsep,leftmargin=20pt]
  \item \textbf{Label-Free Anomaly Detection:} Requires only normal (non-anomalous) data for training by leveraging a leave-one-feature-out strategy that treats each feature as a pseudo-label, eliminating the need for labeled anomalies.
  
  \item \textbf{Context-Aware Scoring:} Captures feature-specific conditional dependencies to accurately reconstruct expected values, enabling precise detection of deviations from normal patterns.
  
  \item \textbf{Interpretability:} Generates cell-level anomaly scores that localize abnormal behavior to specific features within a sample, supporting transparent diagnosis and root cause analysis.
  
  \item \textbf{Scalability:} Naturally supports distributed training and highly parallel inference, allowing the framework to scale efficiently to large, high-dimensional tabular datasets.
\end{itemize}

\begin{figure*}[t]
  \centering
  \includegraphics[width=0.99\textwidth]{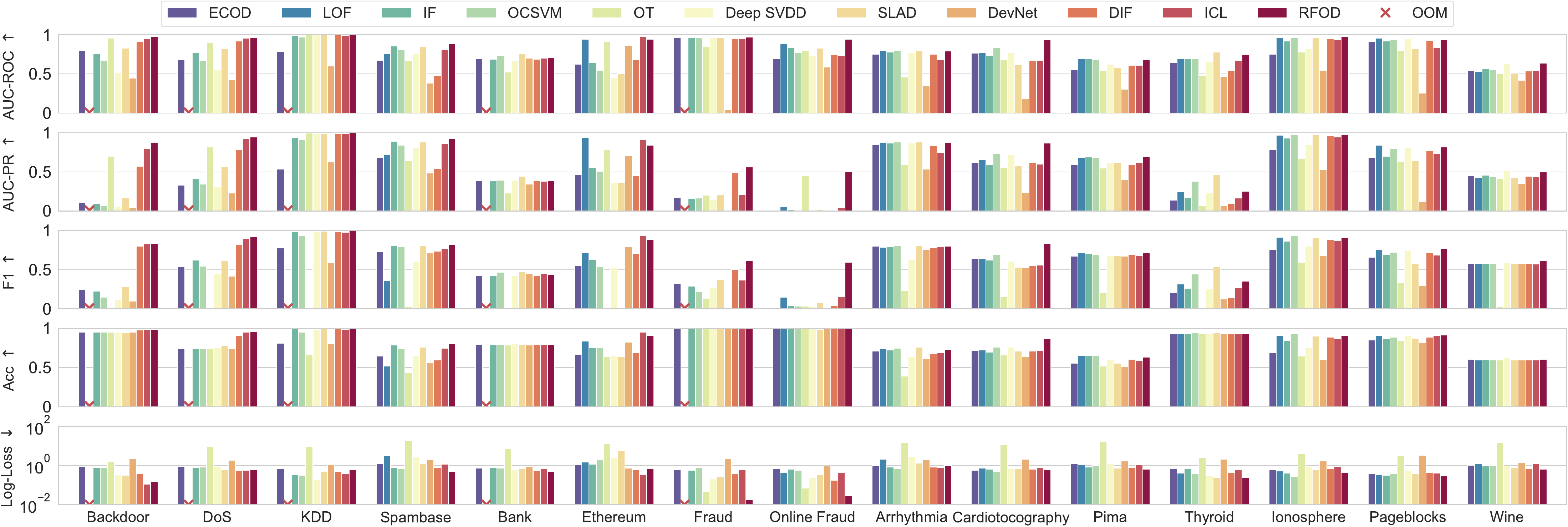}
  \vspace{-1.0em}
  \caption{Detection accuracy of all evaluated methods across 15 benchmark tabular datasets.}
  \label{fig:main_results_all}
  \vspace{-0.75em}
\end{figure*}

\section{Experiments}
\label{sect:expt}

We conduct extensive experiments to rigorously assess the effectiveness and practical utility of \textsf{RFOD} across diverse real-world tabular datasets, guided by the following key research questions:
\begin{itemize}[nolistsep,leftmargin=20pt]
  \item \textbf{Detection Accuracy:} Does \textsf{RFOD} achieve superior detection accuracy compared to state-of-the-art baselines? (\sec{\ref{sect:expt:result}})

  \item \textbf{Efficiency:} Can \textsf{RFOD} perform outlier detection in a time-efficient manner suitable for practical use? (\sec{\ref{sect:expt:efficiency}})

  \item \textbf{Ablation Study:} How essential are the individual modules of \textsf{RFOD}, such as Forest Pruning, Adjusted Gower's Distance, and Uncertainty Weighting? (\sec{\ref{sect:expt:ablation}})

  \item \textbf{Interpretability:} Do the cell-level and row-level scoring mechanisms of \textsf{RFOD} provide meaningful and actionable explanations for detected anomalies? (\sec{\ref{sect:expt:case}})
  
  \item \textbf{Scalability:} How does \textsf{RFOD} scale with increasing dataset sizes in terms of both accuracy and runtime? (\sec{\ref{sect:expt:scalability}})
  
  \item \textbf{Parameter Sensitivity:} How sensitive is \textsf{RFOD} to its hyperparameters? (\sec{\ref{sect:expt:hyperparam}}).
\end{itemize}

We first present the experimental setup, followed by results and analysis for each of the above research questions.

\subsection{Experimental Setup}
\label{sect:expt:setup}

\paragraph{Datasets}
We evaluate \textsf{RFOD} on 15 publicly available tabular datasets spanning diverse domains like cybersecurity, finance, and healthcare. 
Table~\ref{tab:benchmark-datasets} summarizes key dataset statistics, including sample counts, feature dimensionality, and anomaly ratios. 
These datasets are widely used in outlier detection research~\cite{survey1,survey2, icl}, providing robust and comprehensive performance evaluation.

\paragraph{Baselines}
We compare \textsf{RFOD} against 10 state-of-the-art methods drawn from both classical data-mining and modern deep-learning paradigms.
The data-mining baselines include \textbf{\textsf{ECOD}} \cite{li2022ecod}, \textbf{\textsf{LOF}} \cite{breunig2000lof}, \textbf{\textsf{IF}} \cite{liu2008if}, \textbf{\textsf{OCSVM}} \cite{wang2019osvm}, and \textbf{\textsf{OT}} \cite{cortes2020explainable}.
The deep-learning baselines comprise \textbf{\textsf{Deep SVDD}} \cite{deepsvdd}, \textbf{\textsf{SLAD}} \cite{slad}, \textbf{\textsf{DevNet}} \cite{devnet}, \textbf{\textsf{DIF}} \cite{dif}, and \textbf{\textsf{ICL}} \cite{icl}. 

\paragraph{Evaluation Measures}
Following standard evaluation protocols \cite{bergman2020classification, zong2018deep, survey1, survey2}, we adopt an unsupervised setting in which 50\% of the normal data is used for training. The remaining normal samples are mixed with all anomaly instances to form the test set. 
Performance is evaluated using:
\begin{itemize}[nolistsep,leftmargin=20pt]
  \item \textbf{AUC-ROC:} Measures the ranking quality of anomaly scores.
  \item \textbf{AUC-PR:} Emphasizes precision and recall, especially informative under severe class imbalance.
  \item \textbf{F1-Score (F1)} and \textbf{Accuracy (Acc):} Threshold-based metrics reflecting classification performance.
  \item \textbf{Log-Loss:} Assesses the calibration of anomaly probabilities.
  \item \textbf{Training Time} and \textbf{Testing Time:} Record wall-clock training and testing times to evaluate efficiency and scalability.
\end{itemize}

\begin{table}[t]
\centering
\small
\setlength\tabcolsep{5pt}
\renewcommand{\arraystretch}{1.1}
\caption{Summary statistics for the 15 benchmark tabular datasets used in our experiments, including sample sizes, feature dimensions, and anomaly ratios.}
\label{tab:benchmark-datasets}
\vspace{-1.0em}
\resizebox{0.99\columnwidth}{!}{
\begin{tabular}{llrrr}
    \toprule
    \rowcolor[HTML]{FFF2CC}
    \textbf{Domain} & \textbf{Dataset} & \textbf{\# Samples}     & \textbf{Feature Dim.} & \textbf{Anomaly Ratio} \\
    \midrule
    & Backdoor \cite{moustafa2015unsw} & 95,329  & 42   & 2.44\% \\
    & DoS \cite{moustafa2015unsw}      & 109,353  & 42   & 14.95\% \\
    & KDD \cite{kdd_cup_1999_data_130} & 4,898,430 & 41 & 19.86\%  \\
    \multirow{-4}{*}{Cybersecurity} & Spambase \cite{spambase_94} & 4,601    & 57   & 39.40\% \\
    \midrule
    \cellcolor[HTML]{DDEBFF} & \cellcolor[HTML]{DDEBFF}Bank \cite{bank_marketing_222}    & \cellcolor[HTML]{DDEBFF}45,211  & \cellcolor[HTML]{DDEBFF}16  & \cellcolor[HTML]{DDEBFF}11.70\% \\
    \cellcolor[HTML]{DDEBFF} & \cellcolor[HTML]{DDEBFF}Ethereum \cite{vagifa_2020_ethereum_frauddetection} & \cellcolor[HTML]{DDEBFF}9,841   & \cellcolor[HTML]{DDEBFF}47  & \cellcolor[HTML]{DDEBFF}22.14\% \\
    \cellcolor[HTML]{DDEBFF} & \cellcolor[HTML]{DDEBFF}Fraud \cite{dal2015creditcardfraud}    & \cellcolor[HTML]{DDEBFF}284,807  & \cellcolor[HTML]{DDEBFF}30  & \cellcolor[HTML]{DDEBFF}0.17\% \\
    \multirow{-4}{*}{\cellcolor[HTML]{DDEBFF}Finance} & \cellcolor[HTML]{DDEBFF}Online Fraud & \cellcolor[HTML]{DDEBFF}80,000 & \cellcolor[HTML]{DDEBFF}6 & \cellcolor[HTML]{DDEBFF}0.14\% \\
    \midrule
    & Arrhythmia \cite{arrhythmia_5}   & 452   & 279 & 45.80\% \\
    & Cardiotocography \cite{cardiotocography_193} & 2,126 & 22  & 22.15\% \\
    & Pima \cite{smith1988using}       & 768   & 8   & 34.90\% \\
    \multirow{-4}{*}{Healthcare}  & Thyroid \cite{thyroid_disease_102} & 6,916  & 21  & 3.61\% \\
    \midrule
    \cellcolor[HTML]{DDEBFF} & \cellcolor[HTML]{DDEBFF}Ionosphere \cite{ionosphere_52} & \cellcolor[HTML]{DDEBFF}351    & \cellcolor[HTML]{DDEBFF}34    & \cellcolor[HTML]{DDEBFF}35.90\% \\
    \cellcolor[HTML]{DDEBFF} & \cellcolor[HTML]{DDEBFF}Pageblocks \cite{page_blocks_classification_78} & \cellcolor[HTML]{DDEBFF}5,473  & \cellcolor[HTML]{DDEBFF}10    & \cellcolor[HTML]{DDEBFF}10.23\% \\
    \multirow{-3}{*}{\cellcolor[HTML]{DDEBFF}Others} & \cellcolor[HTML]{DDEBFF}Wine \cite{wine_109}  & \cellcolor[HTML]{DDEBFF}4,898  & \cellcolor[HTML]{DDEBFF}11    & \cellcolor[HTML]{DDEBFF}25.38\% \\    
    \bottomrule
\end{tabular}}
\end{table}

\paragraph{Parameter Settings}
For all baselines, we adopt hyperparameter settings recommended in their original papers~\cite{li2022ecod, breunig2000lof, liu2008if, wang2019osvm, cortes2020explainable, deepsvdd, slad, devnet, dif, icl}. 
To ensure fair comparisons, we standardize the number of training epochs to 50 for deep models. 
Moreover, for models that require a contamination parameter, we set it equal to the true anomaly ratio in each dataset. 
Additionally, for \textsf{Deep SVDD}, \texttt{n\_features} matches each dataset's feature dimensionality.

\paragraph{Environment}
All experiments are conducted on a workstation equipped with an Intel\textsuperscript{\textregistered} Xeon\textsuperscript{\textregistered} Platinum 8480C @3.80GHz, 256 GB of RAM, and an NVIDIA H200 GPU.

\begin{figure*}[t]
  \centering
  \includegraphics[width=0.99\textwidth]{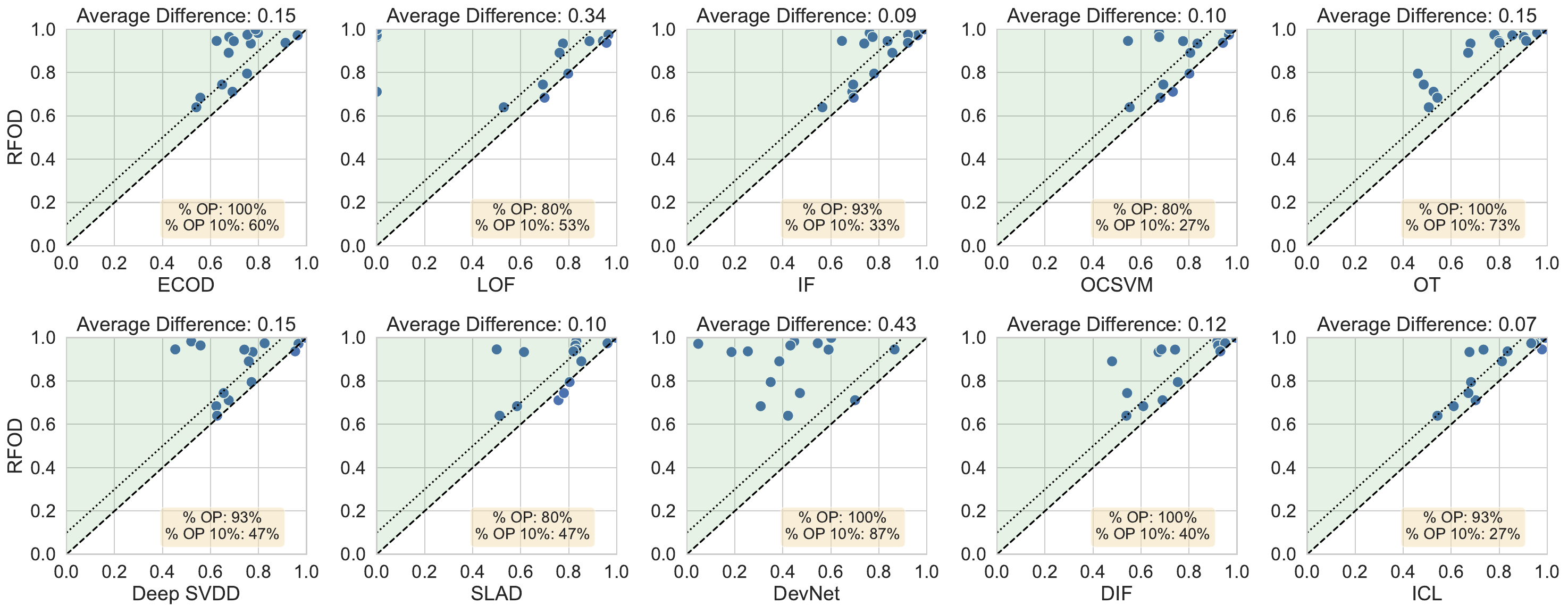}
  \vspace{-1.0em}
  \caption{Pairwise comparisons on AUC-ROC between \textsf{RFOD} and other competitors.}
  \label{fig:pairwise}
  \vspace{-0.5em}
\end{figure*}

\subsection{Detection Accuracy}
\label{sect:expt:result}

\paragraph{Main Results}
Figure \ref{fig:main_results_all} summarizes the performance of all methods across the 15 benchmark datasets. 
%%%
Although \textsf{RFOD} is occasionally outperformed on specific datasets, it consistently secures either the highest or second-highest scores across all metrics and baselines. 
This reflects not only its strong peak performance but also its overall reliability.
%%%
In particular, \textsf{RFOD} exhibits substantial improvements in AUC-PR, a critical metric for \emph{imbalanced} outlier detection.
%%%
Compared to data mining-based methods, \textsf{RFOD} achieves an average AUC-PR gain of 46.7\%, and it surpasses deep learning-based approaches by 24.8\% in AUC-ROC.
These results demonstrate \textsf{RFOD}'s capacity to accurately identify true anomalies while maintaining low false positive rates, which is especially crucial in practical scenarios where anomalies are rare yet costly to overlook.

On datasets where \textsf{RFOD} does not rank first, it is outperformed by diverse competitors such as \textsf{OCSVM}, \textsf{LOF}, \textsf{DIF}, and \textsf{SLAD}, rather than by a single consistently superior method.
%%%
This suggests that no existing alternative uniformly outperforms \textsf{RFOD}, further underscoring its broad applicability across various data characteristics.
%%%

\paragraph{Rankings Analysis}
Table~\ref{tab:avg_rank} presents the average rankings of all methods across five evaluation metrics, where lower values indicate better performance.
%%%
While \textsf{RFOD} might not always achieve first place on every individual dataset, its persistently high rankings highlight its robustness and excellent generalization capabilities.
%%%
Collectively, these demonstrate that \textsf{RFOD} excels not only in achieving superior performance on specific benchmarks but also in delivering stable, reliable results across diverse scenarios.

\begin{table}[ht]
\small
\centering
\renewcommand{\arraystretch}{1.0}
\caption{Average ranks of all evaluated methods computed over 15 real-world tabular datasets. Lower ranks indicate better overall performance across the benchmarks.}
\label{tab:avg_rank}
\vspace{-1.0em}
\resizebox{0.99\columnwidth}{!}{
\begin{tabular}{lrrrrr}
    \toprule
    \cellcolor[HTML]{FFF2CC} & \multicolumn{5}{c}{\cellcolor[HTML]{FFF2CC}\textbf{Average Rank $\downarrow$}} \\ \cmidrule(lr){2-6}
    \cellcolor[HTML]{FFF2CC}\multirow{-2.5}{*}{\textbf{Method}} & \cellcolor[HTML]{FFF2CC}\textbf{AUC-ROC} & \cellcolor[HTML]{FFF2CC}\textbf{AUC-PR} & \cellcolor[HTML]{FFF2CC}\textbf{F1} & \cellcolor[HTML]{FFF2CC}\textbf{Acc} & \cellcolor[HTML]{FFF2CC}\textbf{Log-Loss} \\
    \midrule
    \textsf{ECOD} \cite{li2022ecod}  & 7.40  & 7.47  & 7.00  & 6.77  & 6.93 \\
    \textsf{LOF} \cite{breunig2000lof}  & 6.00  & 6.27  & 6.63  & 6.50  & 4.73 \\
    \textsf{IF} \cite{liu2008if}   & 5.13  & 5.73  & 5.07  & 5.53  & 5.47 \\
    \textsf{OCSVM} \cite{wang2019osvm} & \underline{4.67}  & \underline{5.00}  & \underline{4.47}  & \underline{5.20}  & 5.20 \\
    \textsf{OT} \cite{cortes2020explainable}   & 7.53  & 7.27  & 10.30 & 8.83  & 9.73 \\
    \textsf{Deep SVDD} \cite{deepsvdd} & 6.13  & 6.00  & 7.13  & 5.20  & 6.40 \\
    \textsf{SLAD} \cite{slad} & 5.20  & 5.07  & 4.77  & 6.00  & 5.20 \\
    \textsf{DevNet} \cite{devnet} & 9.93  & 10.07 & 8.57  & 8.77  & 9.60 \\
    \textsf{DIF} \cite{dif}  & 6.53  & 6.13  & 5.40  & 5.60  & \underline{4.27} \\
    \textsf{ICL} \cite{icl}  & 5.67  & 5.07  & 4.87  & 5.27  & 5.73 \\
    \rowcolor[HTML]{D9EAD3} 
    \textsf{RFOD}  & \textbf{1.80} & \textbf{1.93} & \textbf{1.80} & \textbf{2.33} & \textbf{2.73} \\
    \bottomrule
\end{tabular}}
\end{table}

\paragraph{Pairwise Comparisons}
To provide finer-grained insights, we conduct pairwise comparisons between \textsf{RFOD} and each baseline. 
%%%
Figure~\ref{fig:pairwise} illustrates scatter plots where each point represents a dataset: the x-axis indicates the AUC-ROC score of the baseline, while the y-axis shows the corresponding score of \textsf{RFOD}.
The solid diagonal line represents parity, and dashed lines highlight a +10\% margin. 
%%%
Each subplot reports three key metrics:
(1) \textbf{Average Difference}: Mean absolute difference in AUC-ROC between \textsf{RFOD} and the baseline.
(2) \textbf{\% OP (Outperformed)}: Percentage of datasets where \textsf{RFOD} outperforms the baseline. 
(3) \textbf{\% OP 10\%}: Percentage of datasets where \textsf{RFOD} exceeds the baseline by at least 10\%. 
%%%
These measures provide complementary views on both the magnitude and consistency of performance gains.

\textsf{RFOD} consistently outperforms both data mining and deep learning baselines on most datasets.
%%%
Compared to data-mining-based approaches, \textsf{RFOD} achieves higher AUC-ROC scores on nearly all benchmarks, with average improvements ranging from 0.07 to 0.43. 
For example, compared to \textsf{OT}, \textsf{RFOD} secures an average AUC-ROC gain of 0.15, exceeding it by more than 10\% on over half the evaluated datasets.
%%%
%%%
Similarly, \textsf{RFOD} demonstrates significant advantages over deep learning methods. Notably, it outperforms \textsf{DevNet} on all datasets, achieving an impressive average improvement of 0.43 in AUC-ROC. 
%%% 
These gains primarily stem from \textsf{RFOD}'s feature-specific modeling and robust handling of mixed-type tabular data, areas where traditional methods often fall short.

\begin{figure*}[t]
  \centering
  \includegraphics[width=0.99\textwidth]{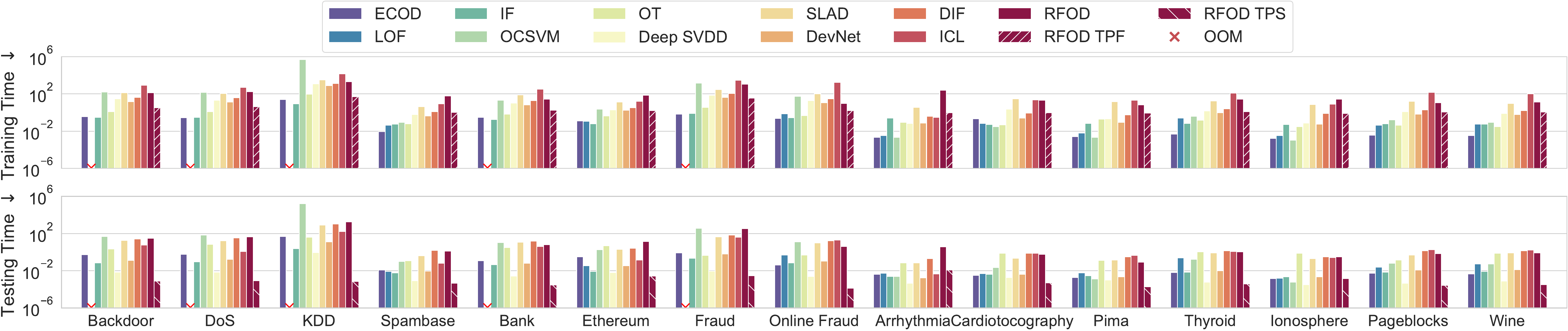}
  \vspace{-1.0em}
  \caption{Training and testing time (seconds) of all evaluated methods across 15 benchmark tabular datasets.}
  \label{fig:efficiency}
  \vspace{-0.75em}
\end{figure*}

\begin{figure*}[t]
  \centering
  \includegraphics[width=0.99\textwidth]{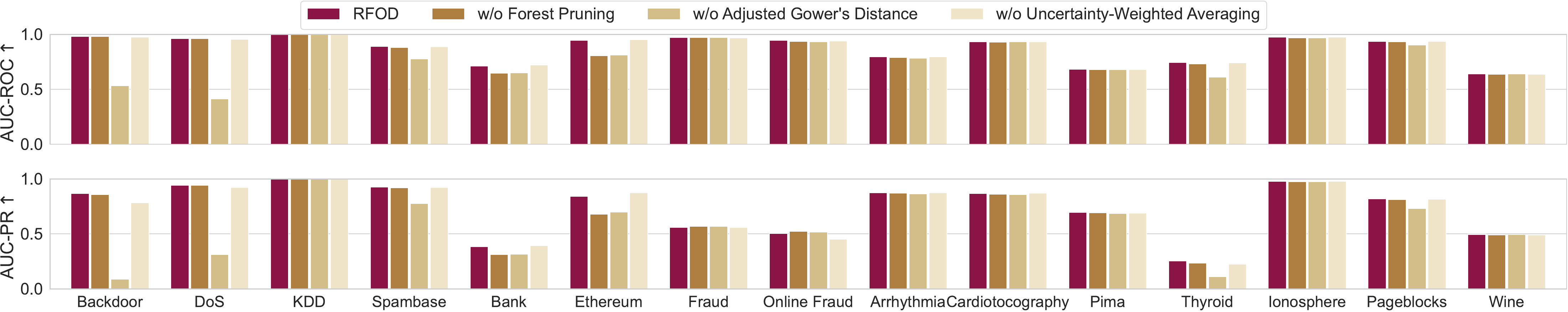}
  \vspace{-1.0em}
  \caption{Ablation study results of \textsf{RFOD} on AUC-ROC and AUC-PR across 15 benchmark tabular datasets.}
  \label{fig:ablation}
  \vspace{-0.5em}
\end{figure*}

\subsection{Efficiency}
\label{sect:expt:efficiency}

We then evaluate both training and inference times in Figure~\ref{fig:efficiency}.

\paragraph{Training Time}
%%%
Data mining-based methods generally exhibit fast training due to their lightweight computations and simple algorithmic designs. 
However, \textsf{LOF} suffers from out-of-memory errors on large datasets such as DoS and Fraud, highlighting its limited scalability for high-volume data.
%%%
In contrast, deep learning-based methods are considerably more time-consuming, requiring multiple epochs and significant computational resources, especially for high-dimensional datasets where training time grows substantially.

%%%
\textsf{RFOD} strikes a balance between efficiency and performance. 
While its training time is slightly higher than most data-mining approaches due to training multiple feature-specific forests, it remains significantly faster than deep-learning alternatives.
%%%
To examine \textsf{RFOD}'s internal efficiency, we decompose its training time into the mean Time Per Feature (TPF), shown as hatched bars in Figure~\ref{fig:efficiency}.
%%%
Since \textsf{RFOD} trains a separate Random Forest for each feature, TPF offers insight into the per-feature computational cost. 
Results show that TPF remains moderate and scalable, indicating that \textsf{RFOD} can effectively support parallel processing, as elaborated in \sec{\ref{sect:methods:algo}}.

\paragraph{Testing Time}
Inference times are broadly comparable across all methods on each dataset.
Notably, \textsf{RFOD} achieves competitive inference speeds, often surpassing data-mining baselines like \textsf{OCSVM} and \textsf{IF} on larger datasets. 
This stems from its forest pruning strategy and streamlined prediction traversal.
%%%
To provide finer detail on \textsf{RFOD}'s inference cost, we report the Time Per Sample (TPS) as hatched bars in Figure~\ref{fig:efficiency}. 
TPS measures the average time required to compute anomaly scores for a single test sample. 
The results show that TPS stays consistently low and stable across datasets, underscoring the practical efficiency of \textsf{RFOD}'s inference pipeline.

\paragraph{Summary}
Overall, \textsf{RFOD} offers an excellent trade-off between detection accuracy and computational efficiency.
%%%
During training, its modular architecture facilitates independent construction of feature-specific forests, enabling straightforward parallelization and reduced wall-clock times.
%%%
At inference, pruning less-informative trees streamlines computation, delivering low-latency predictions even for high-dimensional datasets.
%%%
Compared to deep-learning baselines, which often incur significant training costs, \textsf{RFOD} presents a practical, resource-efficient solution for large-scale anomaly detection in real-world tabular data scenarios.

\subsection{Ablation Study}
\label{sect:expt:ablation}

To quantify the contribution of each core component in \textsf{RFOD}, we perform an ablation study by individually disabling three key modules: (1) Forest Pruning, (2) AGD for cell-level scoring, and (3) UWA for row-level aggregation.
%%%
Each variant alters only one component while keeping the rest of the architecture intact:
\textbf{\textsf{w/o Forest Pruning}} retains all trees without OOB-based pruning; 
\textbf{\textsf{w/o Adjusted Gower's Distance}} replaces AGD with standard Gower's Distance for cell-level scoring;
and \textbf{\textsf{w/o Uncertainty-Weighted Averaging}} uses simple averaging instead of UWA for aggregating cell-level scores.
%%%
We evaluate these variants on all 15 benchmark datasets, reporting results for AUC-ROC and AUC-PR in Figure~\ref{fig:ablation}. 
Similar trends hold for F1-score, Accuracy, and Log-Loss.

\paragraph{Effect of Forest Pruning} 
Forest Pruning consistently improves performance by removing low-quality trees based on OOB validation.
%%%
On datasets such as Bank, Ethereum, and Thyroid, where overfitting is common due to small sample sizes or high dimensionality, pruning yields clear gains in both AUC-ROC and AUC-PR. 
%%%
This selective retention not only reduces computation but also enhances generalization by eliminating noisy decision paths.

\paragraph{Effect of Adjusted Gower's Distance}
Among all components, AGD proves most critical.
Replacing AGD with standard GD results in the steepest performance drops across nearly all datasets.
%%%
For instance, on the \textsf{DoS} dataset, AUC-ROC plunges from 0.96 to 0.41, % and AUC-PR falls from 0.94 to 0.31, 
a reduction of over 50\% in this metric.
%%%
Similar degradations occur in skewed or heavy-tailed datasets like Spambase, Thyroid, and Pageblocks, underscoring AGD's role in robust scaling and confidence-aware matching for detecting subtle anomalies in mixed-type data.

\paragraph{Effect of Uncertainty-Weighted Averaging} 
Disabling UWA and using unweighted averaging causes moderate yet consistent performance drops, especially on large datasets like Backdoor and DoS.
%%%
Without UWA, noisy tree predictions overly influence row-level scores.
UWA mitigates this by down-weighting high-variance predictions and amplifying reliable signals, resulting in more stable and precise anomaly rankings.

\paragraph{Summary} 
Overall, the full \textsf{RFOD} framework, with all components enabled, delivers the highest AUC-ROC and AUC-PR across nearly every dataset.
%%%
The performance degradation observed when removing any single module confirms the complementary roles of Forest Pruning, AGD, and UWA. 
%%%
These findings validate the architectural choices behind \textsf{RFOD} and demonstrate that its superior performance stems from the synergistic integration of all three components rather than reliance on any individual part.

\begin{figure}[t]
\centering
\subfigure[Comparisons of anomaly scoring.]{
  \label{fig:casestudy_pima:heatmap}
  \includegraphics[width=0.45\columnwidth]{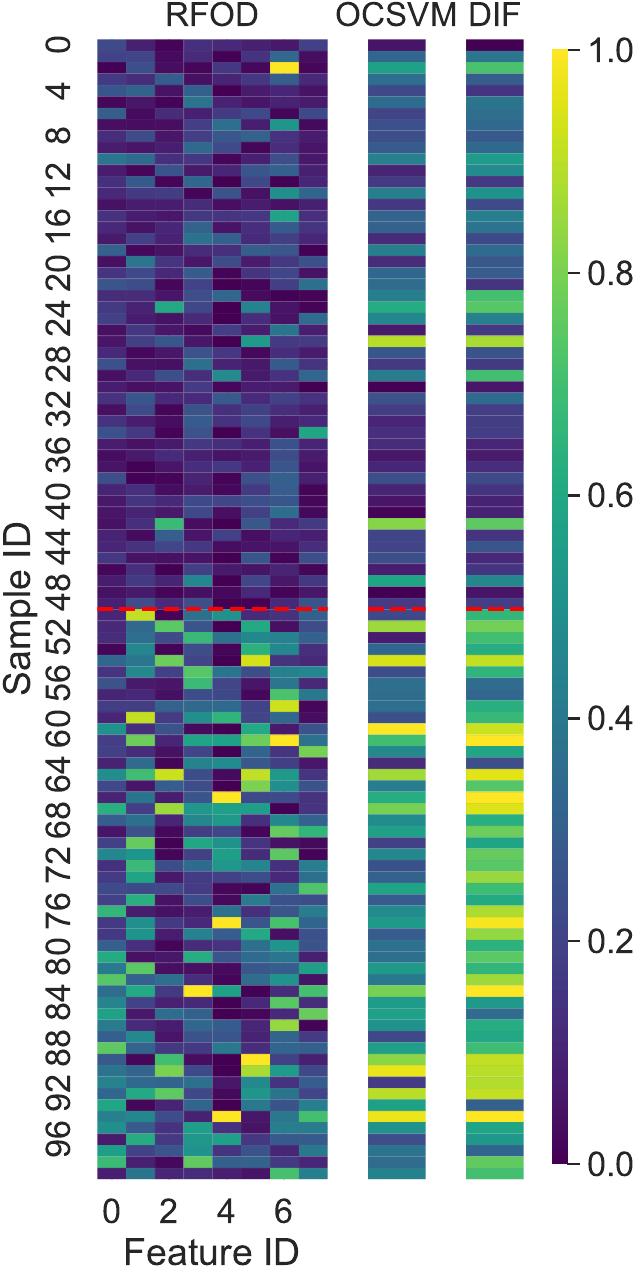}}
\subfigure[Analysis of one outlier on \textsf{RFOD}.]{
  \label{fig:casestudy_pima:analysis}
  \includegraphics[width=0.47\columnwidth]{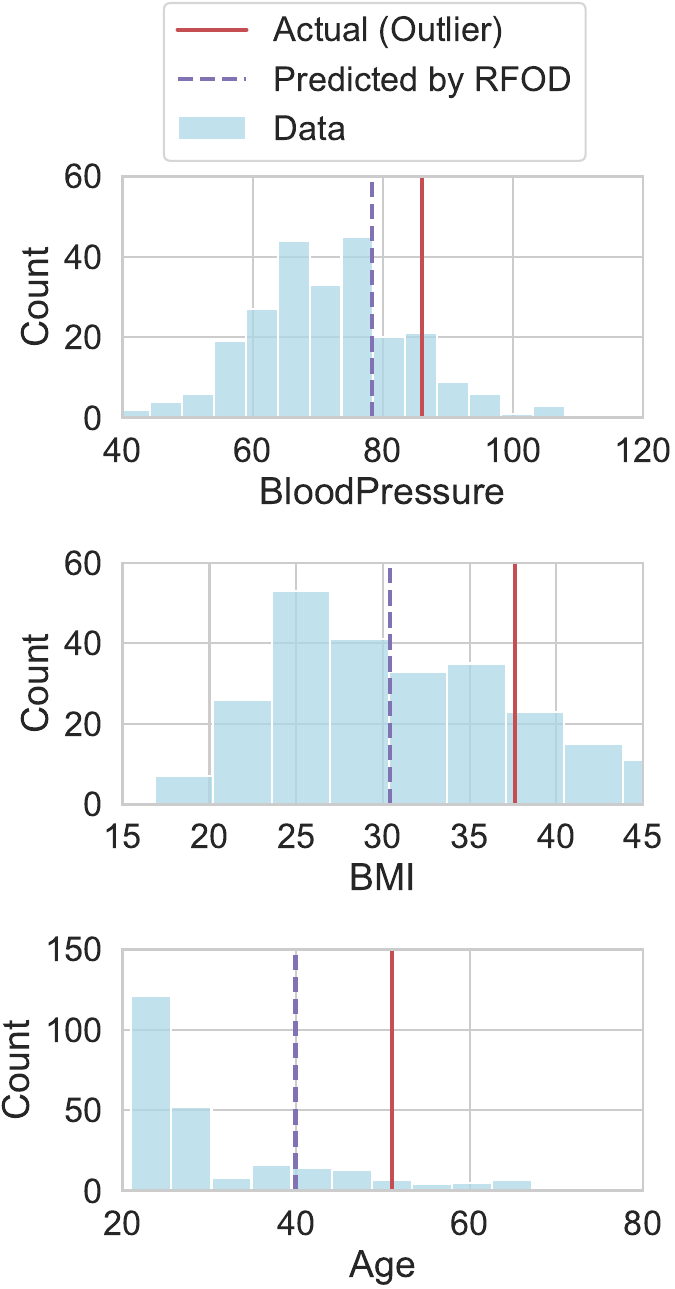}}
\vspace{-1.25em}
\caption{Case study on Pima (Healthcare).}
\label{fig:casestudy_pima}
\vspace{-1.0em}
\end{figure}

\begin{figure}[t]
\centering
\subfigure[Comparisons of anomaly scoring.]{
  \label{fig:casestudy_dos:heatmap}
  \includegraphics[width=0.45\columnwidth]{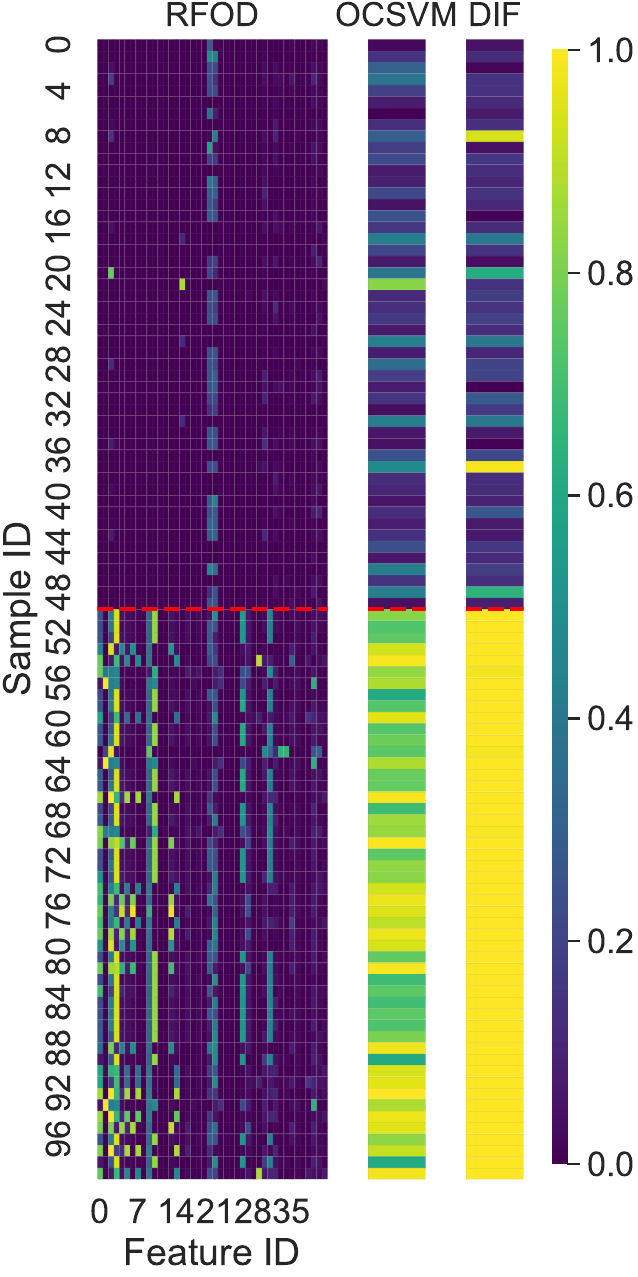}}
\subfigure[Analysis of one outlier on \textsf{RFOD}.]{
  \label{fig:casestudy_dos:analysis}
  \includegraphics[width=0.47\columnwidth]{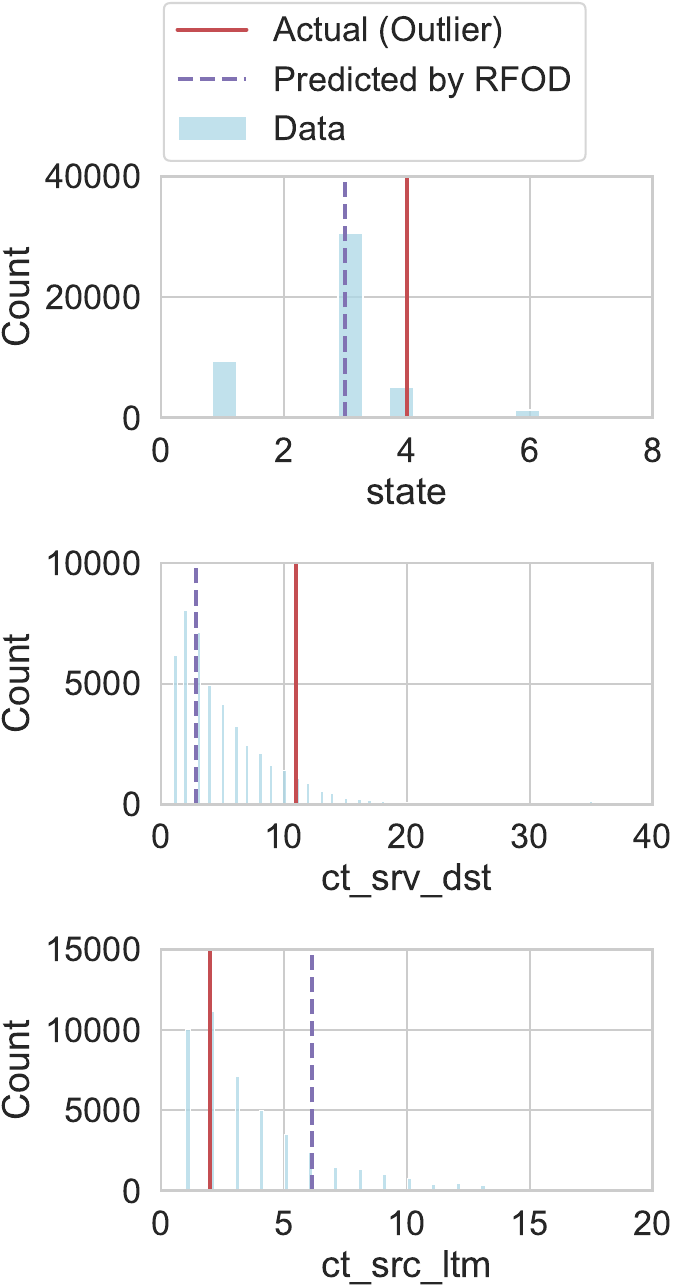}}
\vspace{-1.25em}
\caption{Case study on DoS (Cybersecurity).}
\label{fig:casestudy_dos}
\vspace{-1.25em}
\end{figure}

\begin{figure}[t]
\centering
\subfigure[Comparisons of anomaly scoring.]{
  \label{fig:casestudy_bank:heatmap}
  \includegraphics[width=0.45\columnwidth]{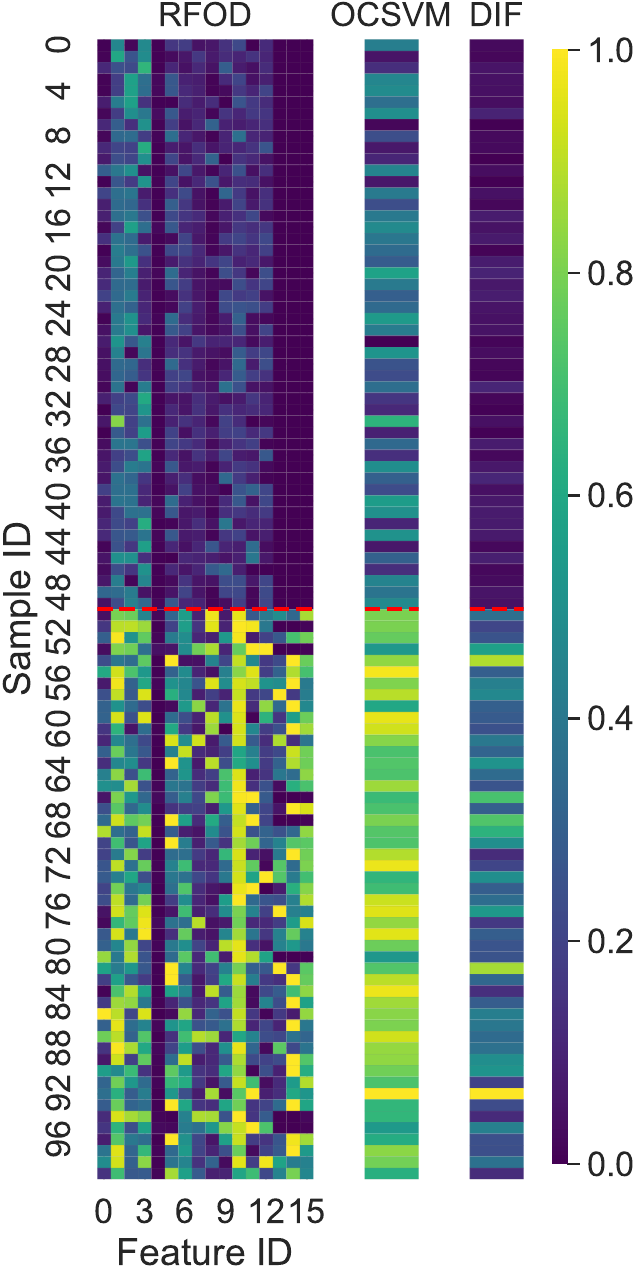}}
\subfigure[Analysis of one outlier on \textsf{RFOD}.]{
  \label{fig:casestudy_bank:analysis}
  \includegraphics[width=0.47\columnwidth]{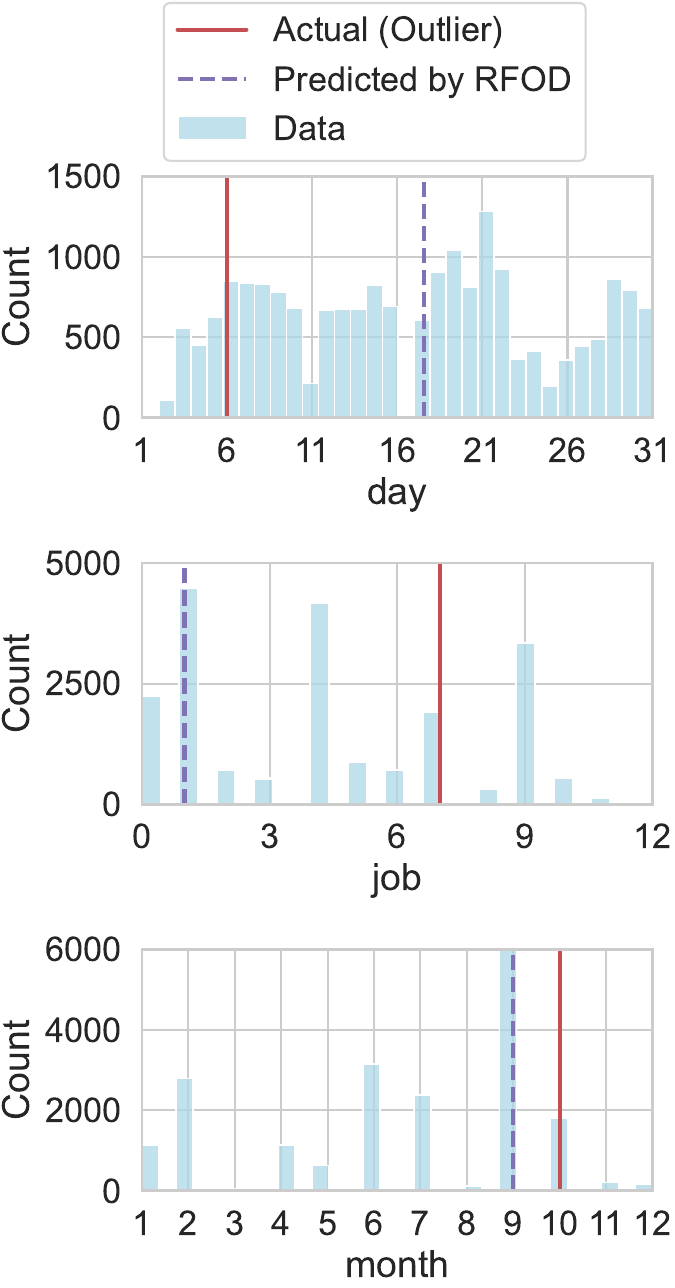}}
\vspace{-1.25em}
\caption{Case study on Bank (Finance).}
\label{fig:casestudy_bank}
\vspace{-1.0em}
\end{figure}

\subsection{Case Study}
\label{sect:expt:case}

To showcase the interpretability and diagnostic capabilities of \textsf{RFOD}, we present three case studies across healthcare, cybersecurity, and finance, as shown in Figures~\ref{fig:casestudy_pima}–\ref{fig:casestudy_bank}.
%%%
Below, we focus on the Pima dataset from the healthcare domain to provide a detailed walkthrough, while similar observations hold for the other datasets.

\paragraph{Comparisons of Cell-Level Anomaly Scores}
The Pima dataset consists of 8 physiological features describing patient health profiles.
Figure~\ref{fig:casestudy_pima:heatmap} displays cell-level anomaly scores for 100 samples across these features using \textsf{RFOD}. 
In this heatmap, darker shades denote lower anomaly scores, while brighter shades highlight potential anomalies, enabling precise, fine-grained insights at the cell level.
%%%
For comparison, the middle and right heatmaps show results from two strong baselines: \textsf{OCSVM} and \textsf{DIF}. Samples 0–49 represent normal instances, while samples 50–99 are labeled as anomalies.
%%%
The visualization shows that normal samples under \textsf{RFOD} typically have sparse, low anomaly scores, while anomalies display concentrated high scores in specific features.
%%%
\textsf{RFOD} highlights distinct feature combinations depending on context, providing targeted explanations for anomalies. 
In contrast, \textsf{OCSVM} and \textsf{DIF} often produce uniformly high activation across entire rows or columns, failing to isolate true anomaly sources and generating false positives even for some normal samples.

\paragraph{Row-Level Interpretability}
Figure~\ref{fig:casestudy_pima:analysis} illustrates \textsf{RFOD}'s row-level interpretability through analysis of a single outlier.
%%%
It shows feature-wise distributions for \texttt{BloodPressure}, \texttt{BMI}, and \texttt{Age}, with \textsf{RFOD}'s predicted values (purple dashed lines) and the outlier's actual values (red solid lines) overlaid.
%%%
The predicted values fall within high-density regions of the normal distribution, reflecting \textsf{RFOD}'s learned expectations, while the actual outlier values lie in the tails, indicating clear anomalies.
%%%
For instance, the patient's blood pressure and BMI are significantly elevated, while age is unusually low given the context.
Such deviations, difficult to spot from marginal histograms alone, are effectively captured by \textsf{RFOD}'s conditional modeling.

\paragraph{Summary}
Overall, this case study illustrates how \textsf{RFOD} bridges the gap between high detection performance and actionable interpretability. 
Its cell-level heatmaps facilitate global feature inspection, while per-instance analyses deliver clear, localized explanations. 
These capabilities are particularly crucial in domains such as healthcare and finance, where understanding the reasons behind an anomaly is as critical as detecting it.

\subsection{Scalability}
\label{sect:expt:scalability}

We assess the scalability of \textsf{RFOD} using the largest dataset, KDD, by creating six dataset variants with progressively larger training sets. 
This evaluation focuses on how detection effectiveness and runtime evolve as the data volume increases.

\begin{figure}[t]
  \centering
  \includegraphics[width=0.99\columnwidth]{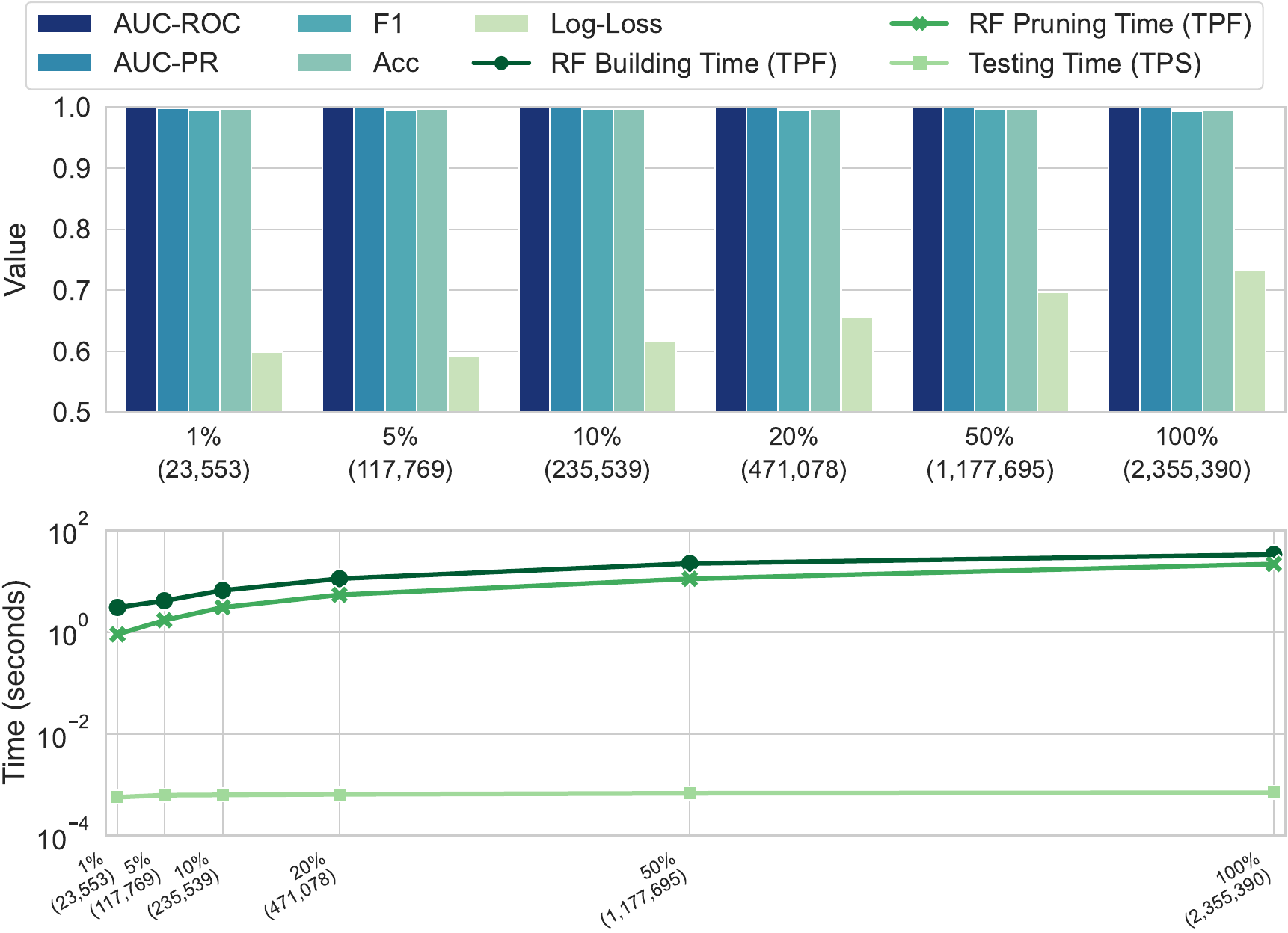}
  \vspace{-1.0em}
  \caption{Scalability results on the KDD dataset.}
  \label{fig:scalability}
  \vspace{-0.75em}
\end{figure}

\paragraph{Effectiveness}
The top panel of Figure~\ref{fig:scalability} shows that \textsf{RFOD} consistently achieves high detection performance, even when trained on as little as 1\% of the data. 
%%%
This robustness stems from its feature-specific random forests, which effectively model conditional relationships and maintain accuracy under data scarcity.

\paragraph{Efficiency}
The bottom panel of Figure~\ref{fig:scalability} breaks down runtime into Random Forest training, pruning, and testing costs.
%%%
Training time scales roughly linearly with the number of samples, consistent with the complexity analysis in \sec{\ref{sect:methods:algo}}. Notably, testing time remains low and stable regardless of training size, highlighting \textsf{RFOD}’s practicality for large-scale applications.

\subsection{Parameter Study}
\label{sect:expt:hyperparam}

%%% overview
Finally, we examine the sensitivity of \textsf{RFOD} to the hyperparameter $\alpha$, which controls the distribution tails in AGD for numerical features.
%%% setup
We vary $\alpha$ across the set $\{0.001, 0.002, 0.005, 0.01, 0.02, 0.05$, $0.1, 0.2, 0.25\}$ and measure AUC-ROC and AUC-PR on six representative datasets covering diverse domains from Table~\ref{tab:benchmark-datasets}. 
Similar trends are observed across other datasets.

\begin{figure}[t]
  \centering
  \includegraphics[width=0.99\columnwidth]{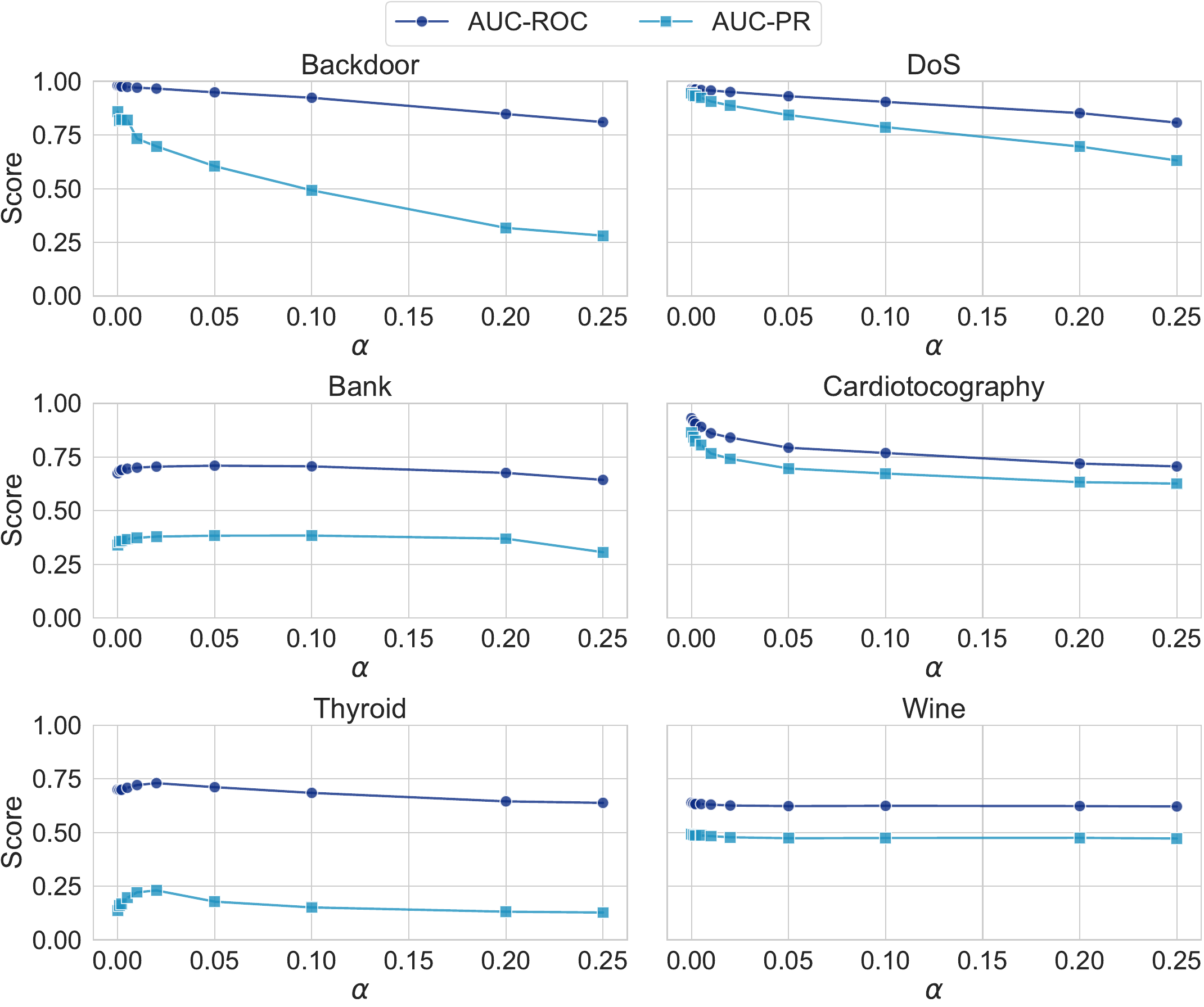}
  \vspace{-1.0em}
  \caption{Impact of varying $\alpha$ on six datasets.}
  \label{fig:para:alpha}
  \vspace{-0.75em}
\end{figure}

%%%
As shown in Figure~\ref{fig:para:alpha}, smaller values of $\alpha$ generally yield higher detection performance.
%%%
Low $\alpha$ values make AGD more sensitive to subtle deviations by focusing on narrow, high-density regions of the data distribution.
%%%
However, excessively small $\alpha$ (e.g., below $0.002$) can introduce numerical instability due to overly tight quantile ranges, offering limited additional benefit.
%%%
Based on these results, we recommend selecting $\alpha$ in the range $[0.01, 0.02]$ for most applications, balancing sensitivity and stability.

\section{Conclusion}
\label{sect:conclusion}

This work introduced \textsf{RFOD}, a novel Random Forest–based framework tailored for outlier detection in tabular data. 
Unlike traditional methods that struggle with heterogeneous features or deep learning approaches that lack interpretability, \textsf{RFOD} reframes anomaly detection as a feature-wise conditional reconstruction task.
%%%
By training dedicated random forests for each feature and leveraging AGD, \textsf{RFOD} delivers precise, cell-level anomaly scoring that remains robust under skewed distributions and categorical uncertainty. 
Together with UWA for row-level aggregation and efficient forest pruning, \textsf{RFOD} strikes an effective balance of detection accuracy, scalability, and interpretability.
%%%.
Extensive experiments across 15 real-world datasets show that \textsf{RFOD} consistently outperforms state-of-the-art baselines in both performance and efficiency while delivering actionable insights into the root causes of anomalies. 
Case studies further demonstrate its practical value in high-stakes domains, highlighting both global and local interpretability.

\balance
\bibliographystyle{ACM-Reference-Format}
\bibliography{main}

\end{document}